\PassOptionsToPackage{numbers, compress}{natbib}
\documentclass{article}
\usepackage[final, main]{neurips_2026}

\usepackage[utf8]{inputenc}
\usepackage[T1]{fontenc}
\usepackage{hyperref}
\usepackage{url}
\usepackage{booktabs}
\usepackage{amsfonts}
\usepackage{nicefrac}
\usepackage{microtype}
\usepackage{amsmath}
\usepackage{amssymb}
\usepackage{mathtools}
\usepackage{amsthm}
\usepackage{graphicx}
\usepackage{subcaption}
\usepackage{colortbl}
\usepackage{multirow}
\usepackage{makecell}
\usepackage[dvipsnames]{xcolor}
\usepackage{bbding}
\usepackage[capitalize,noabbrev]{cleveref}
\usepackage{caption}
\usepackage{titletoc}

\theoremstyle{plain}

\theoremstyle{definition}

\theoremstyle{remark}

\usepackage[textsize=tiny]{todonotes}

\title{Going Down Memory Lane: Scaling Tokens for Video Stream Understanding with Dynamic KV-Cache Memory}

\author{%
  Vatsal Agarwal\thanks{Work partially done during internship at TikTok. This work is for research purposes only and is not currently integrated into any TikTok technology.} \\
  University of Maryland, College Park \\
  \texttt{vatsalag@umd.edu} \\
  \And
  Saksham Suri \\
  University of Maryland, College Park \\
  \And
  Matthew Gwilliam \\
  TikTok \\
  \And
  Pulkit Kumar \\
  University of Maryland, College Park \\
  \And
  Abhinav Shrivastava \\
  University of Maryland, College Park \\
}

\begin{document}
\maketitle

\begin{abstract}

Online video understanding requires models to robustly encode, store, and retrieve information from a continuous stream to support accurate question answering. 
Existing approaches rely on key--value caching to accumulate frame-level representations over time, but allocate only a limited token budget per frame, causing loss of fine-grained visual detail. 
We find that naively scaling this budget harms retrieval quality and question-answering performance, as denser token streams introduce redundancy that degrades query--frame similarity.
To address this, we first introduce an adaptive selection strategy that reduces token redundancy while preserving local spatiotemporal information.
Second, we propose a training-free retrieval ensemble that leverages external models to better identify relevant frames.
Our method, \textbf{MemStream}, achieves +8.0\% on CG-Bench, +8.5\% on LVBench, and +2.4\% on VideoMME (Long) over ReKV with Qwen2.5-VL-7B. 
Furthermore, we show that MemStream consistently improves performance across several Video-LLMs for long-video understanding. 
Project page: \url{https://vatsalag99.github.io/memstream/}

\end{abstract}

\section{Introduction}
\label{sec:introduction}

Video understanding is a complex task requiring a model to perceive and reason about the scene, the objects in it, and temporal interactions such as actions or events.
Recent developments in multimodal large language models (MLLMs) have equipped them with the capabilities to understand these complex relations~\cite{liu2023visual,li2024llava,bai2023qwen,wang2024qwen2,yang2025qwen3,zhu2025internvl3,team2025gemma,he2024malmmmemoryaugmentedlargemultimodal,cheng2024videollama2advancingspatialtemporal}.
However, processing long videos remains a significant challenge due to limited context length.
Capturing fine-grained visual details requires the model to encode and retain information at both high spatial and temporal granularity, yet current models must sacrifice this ability to fit within the context budget. 

Existing approaches use a combination of temporal subsampling, where a subset of key frames is selected from the video~\cite{liu2025bolt,tang2025adaptive,yao2025generativeframesamplerlong}, or spatial subsampling~\cite{cheng2024videollama2advancingspatialtemporal,bai2025qwen2,wang2024qwen2}, where a subset of tokens are processed for each frame.
However, sampling only a sparse subset of frames results in the model missing granular temporal information such as short but critical events~\cite{di2025rekv,sun2025framesclipstrainingfreeadaptive}, while low frame-wise token budgets can result in the model missing fine-grained visual details~\cite{nie2024slowfocus}. 
 Recent dynamic-resolution MLLMs such as Qwen2.5-VL~\cite{bai2025qwen2} make it possible to process frames at higher spatial token density; however, doing so across long videos still causes it to exceed the context budget. 

A natural way to address these limitations is by processing the video in a streaming manner, where frames are processed as they arrive and stored incrementally for question-answering. This is exemplified by ReKV~\cite{di2025rekv}, which encodes video via causal sliding-window attention and stores the resulting representations directly in the model's key--value (KV) cache. During question-answering, relevant frames are retrieved from this cache across each layer. In this way, the model no longer needs to re-encode the video each time for a new question and can process the video beyond its original context length. Moreover, the KV-cache can serve as an efficient memory that can be queried to answer multiple questions. Crucially, this design couples encoding, retrieval, and question-answering through a single set of KV representations, making model performance fundamentally dependent on their quality. If these cache representations are not discriminative or fail to capture fine-grained visual information, the model may fail to retrieve the right visual evidence or produce incorrect answers.

\begin{figure}[t]
\begin{center}
\includegraphics[width=\linewidth]{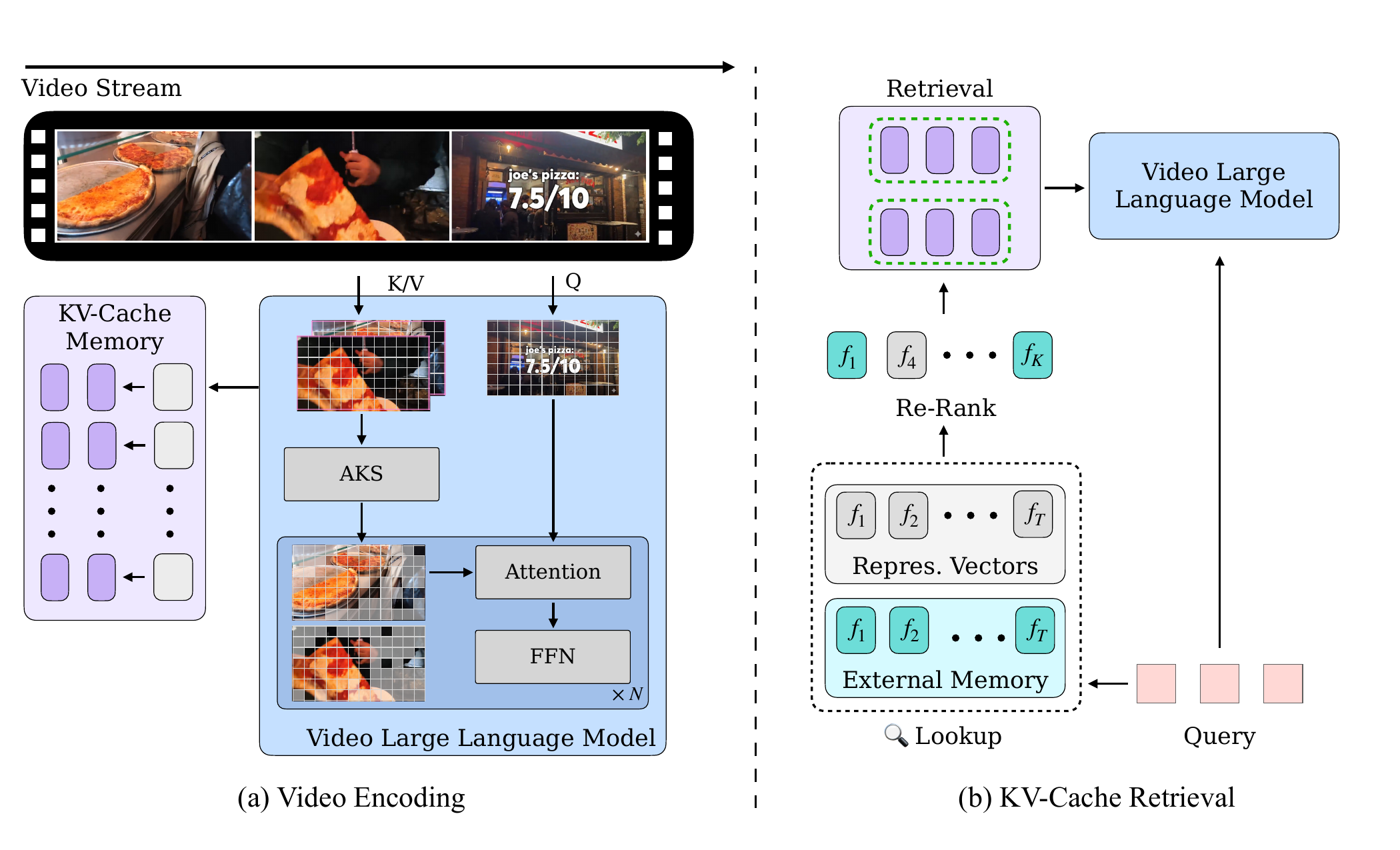}
\caption{(a) We propose constructing the key--value cache with sparse sliding-window attention using our adaptive key selection (AKS) strategy. (b) During question-answering, we merge complementary retrieval signals from external models via a training-free ensemble.}
\label{fig:teaser}
\end{center}
\end{figure}

To better understand this bottleneck, we conduct a detailed analysis of existing KV-cache-based methods. Our exploration reveals two critical limitations of this design. First, we observe that increasing per-frame token budget counterintuitively degrades both retrieval quality and downstream question-answering performance. We identify that this is a result of the sliding-window attention, which we hypothesize amplifies local redundancy within the key--value features, resulting in less discriminative frame-wise representations. Second, we find that retrieval quality is unreliable even at lower token budgets: while some layers consistently identify the relevant video segment, others miss it entirely. Moreover, we observe that internal KV features alone often lack sufficient fine-grained visual detail, particularly for questions involving precise object attributes or subtle motion cues.

Motivated by these findings, we propose MemStream, a training-free framework for online video understanding that addresses both failure modes. First, we introduce Adaptive Key Selection (AKS) during video encoding to preserve informative content in the sliding window while discarding redundant signals.
In this way, AKS produces discriminative KV-cache representations that enhance retrieval fidelity and improve downstream question-answering performance. 
To address variable retrieval quality, we propose a training-free retrieval ensemble that utilizes external models to supplement internal KV features. 
This design leverages complementary retrieval signals across internal and external features, yielding more consistent retrieval across layers and improved overall retrieval quality. 
Furthermore, our retrieval ensemble enables the integration of stronger vision encoders as they become available, without any retraining. 

In summary, our core contributions are:
\textbf{(1)} an extensive analysis revealing that KV-cache-based methods suffer from representational degradation at high token budgets and unreliable retrieval quality;
\textbf{(2)} Adaptive Key Selection (AKS), a sparse sliding-window attention strategy that reduces redundancy while preserving informative spatiotemporal content; and
\textbf{(3)} a training-free retrieval ensemble that combines internal KV-cache retrieval with external vision-language retrieval signals.

\section{Related Work}
\label{sec:related_work}

\paragraph{Long Video Understanding}
Recent advances have endowed Video-LLMs with powerful video understanding capabilities~\cite{bai2023qwen,zhu2025internvl3,yang2025qwen3,bai2025qwen2,li2024llava}. These models typically consist of a vision encoder to process frames~\cite{radford2021learning,zhai2023sigmoid,tschannen2025siglip}, and a large-language model (LLM)~\cite{ouyang2022training,touvron2023llama,bai2023qwen} to generate text responses based on user input~\cite{liu2023visual,bai2025qwen2}. 
Despite this progress, these models still struggle with long-form video due to limited context length, making precise temporal localization and reasoning particularly challenging~\cite{fu2025video,chen2024cg,liu2025timescope}.
To address these issues, existing approaches have primarily focused on designing effective compression strategies via spatial and temporal subsampling~\cite{wang2025dynamic, tao2025dycoke, liu2025hybrid, lee2024video,liu2025video, fan2026flashvid, yao2025timechat, liu2025timescope}.
Recently,~\cite{song2025videonsanativesparseattention} proposed a training-based approach to introduce learned sparsity to existing Video-LLMs, showcasing improved performance on long-form video understanding. While token-compression methods mitigate redundancy, they risk losing critical fine-grained information. 
In contrast, our approach selectively preserves tokens only within the local attention window, maintaining a non-evicting KV-cache memory for long-term storage in a training-free manner. 

\paragraph{Streaming Video Understanding}
Streaming video understanding requires the model to process and store frames as they arrive in an online manner. Then, during question-answering, the stored features are queried across a variety of different types of questions~\cite{di2025rekv, videollm-online, zhang2024flash, he2024malmmmemoryaugmentedlargemultimodal, song2024moviechat}. Recent research has explored many different ways to process and store this visual information to obtain an optimal trade-off between efficiency and accuracy. One key area of interest is designing an effective memory for storing and retrieving frame information~\cite{zhang2025flash, zeng2025streamforest, chen2025streamingtom, di2025rekv, kim2025infinipot}. These methods broadly either store and compress visual information before feeding it to the LLM~\cite{xie2026fluxmem, zeng2025streamforest, zhang2025flash}, or use the LLM's internal KV-cache to store intermediate visual features~\cite{kim2025infinipot, di2025rekv, zhang2026hermeskvcachehierarchical, chen2025streamingtom, yang2025streammemqueryagnostickvcache}. Several works have also explored compression strategies for streaming video~\cite{kim2025infinipot,yao2025timechat,chen2025streamingtom,yang2025streammemqueryagnostickvcache,ning2025livevlmefficientonlinevideo}. These methods either apply compression on tokens before feeding them into the  Video-LLM~\cite{yao2025timechat, chen2025streamingtom}, or apply compression directly on the KV-cache~\cite{yang2025streammemqueryagnostickvcache, kim2025infinipot}. 
While these strategies improve efficiency, they discard discrete frame-level information. This not only loses per-frame detail but also reduces retrieval expressivity: different questions may require different subsets of frames, yet a compressed cache must serve all queries with the same fixed representation, collapsing an exponentially large retrieval space into one.

\section{Preliminaries}
\label{sec:Prelim}

Formally, we define the encoding process as follows.
Let the full video stream be $V^T$ with $T$ total frames.
For each video frame $f_t$ at time $t$, let $Q_t^i$ denote the query feature at layer $i$, and let
$K_t^i \in \mathbb{R}^{N \times D}$ and $V_t^i \in \mathbb{R}^{N \times D}$ denote the corresponding
key and value features, where $N$ is the number of tokens per frame and $D$ is the feature dimension.

Let $\omega$ denote the sliding-window size in frames.
At layer $i$, the sliding window consists of the key--value features from the previous $\omega$ frames, $W_t^i = \{K_j^i, V_j^i\}_{j=t-\omega}^{t-1}$ containing a total of $\omega N$ tokens. We define the concatenated window keys and values as
\[
K_{W_t}^i = \mathrm{concat}(\{K_j^i\}_{j=t-\omega}^{t-1}), \quad V_{W_t}^i = \mathrm{concat}(\{V_j^i\}_{j=t-\omega}^{t-1}).
\]

The representation of frame $f_t$ at layer $i$ is then computed as
\[
O_t^i = \mathrm{Attn}\!\left(
Q_t^i,\;
\big[ K_{W_t}^i \,;\, K_t^i \big],\;
\big[ V_{W_t}^i \,;\, V_t^i \big]
\right).
\]
For the next frame $f_{t+1}$, the key--value features $K_t^i$ and $V_t^i$ are appended to the sliding window $W_{t+1}^i$ and key--value pair $\{K_{t-\omega}^i, V_{t-\omega}^i\}$ is removed and offloaded. At this stage, we average-pool each frame feature to form a representative vector $\mathbf{k}_{t-\omega}^i = \frac{1}{N} \sum_{n=1}^{N} K_{t-\omega}^{i}[n]$. These vectors are stored on GPU. 

During question answering, let $\mathcal{Q}^i \in \mathbb{R}^{N_q \times D}$ denote the question embedding at layer $i$, where $N_q$ is the number of tokens in the question. We compute a question representation $\mathbf{q}^i = \frac{1}{N_q} \sum_{n=1}^{N_q} \mathcal{Q}^i[n]$ for frame retrieval.
Query--frame scores $S_\text{internal} \in \mathbb{R}^{T}$ are computed via cosine similarity between $\mathbf{q}^i$
and the set of representative frame vectors $\{\mathbf{k}_j^i\}_{j=1}^{T}$.

Let $R$ denote the indices of the top-$k$ retrieved frames.
The corresponding key--value pairs are then retrieved for question answering.
Using the retrieved keys $K_R^i$ and values $V_R^i$, the output is computed as
\[
O^i = \mathrm{Attn}\!\left(
Q^i,\;
\big[ K_R^i \,;\, K^i \big],\;
\big[ V_R^i \,;\, V^i \big]
\right).
\]

\section{Analysis}
\label{sec:Analysis}

In this section, we analyze how scaling per-frame token budgets affects model performance for dynamic resolution models. Notably, we find that naively scaling the per-frame token budget counterintuitively degrades both 
retrieval quality and downstream question-answering performance in KV-cache-based methods, despite the intuition that richer visual representations should improve understanding.

\subsection{Scaling Token Budget Hurts Performance}
\label{subsec:scaling}

Prior work on KV-cache compression has largely been applied to models with static token budgets, such as LLaVA-OneVision or LLaVA-Video.
In contrast, few studies integrate KV-cache memory for dynamic-resolution models such as Qwen2.5-VL~\cite{zhang2026hermeskvcachehierarchical}.
We integrate Qwen2.5-VL with KV-cache encoding and retrieval, and apply it to CG-Bench, which provides the "clue" frames required to answer the question.
This allows us to examine not only whether the model answers the question correctly, but also whether the model retrieves relevant frames from its KV-cache.
All experiments use 128 input frames. Since Qwen2.5-VL represents two frames with one feature, this is equivalent to 64 frame-wise features.

\begin{figure*}[t]
\begin{minipage}{0.48\linewidth}
    \centering
    \includegraphics[width=\linewidth]{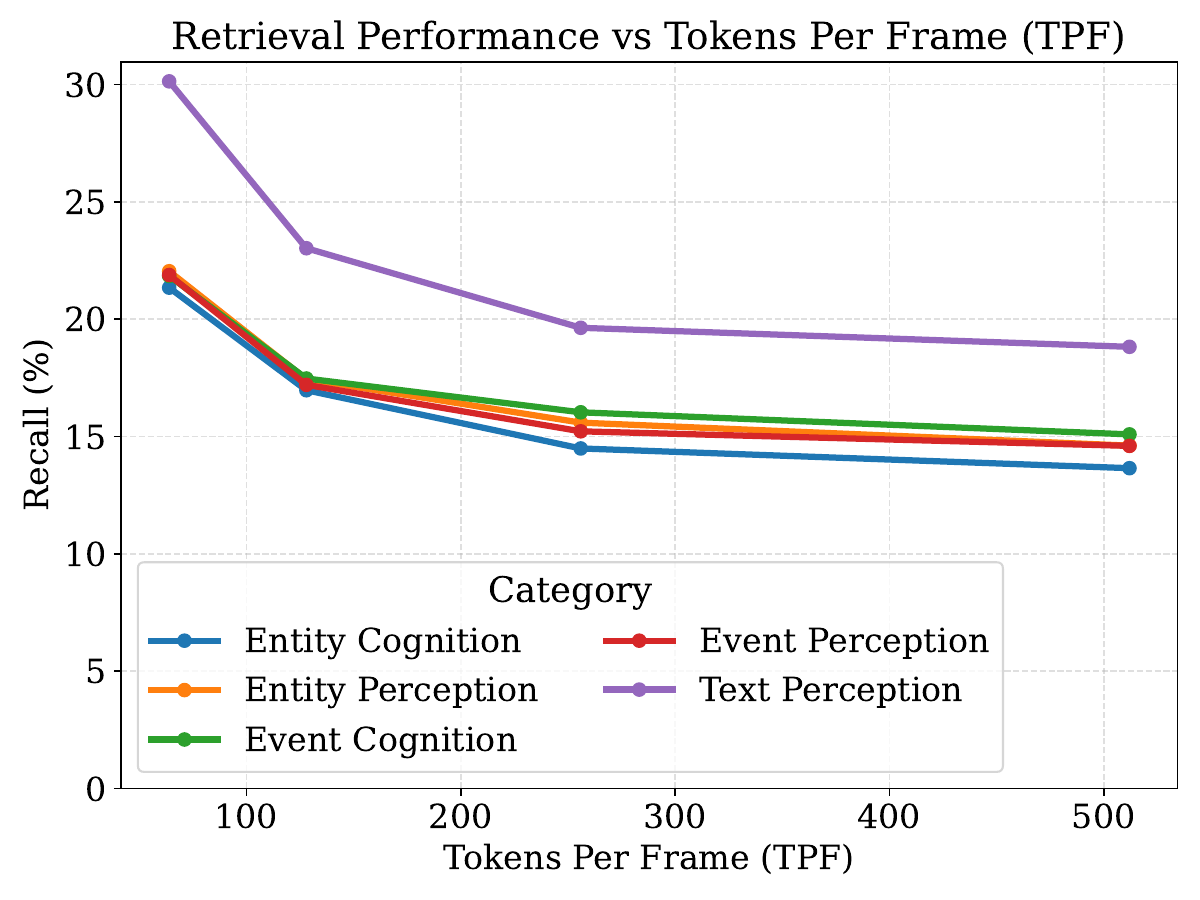}
    \caption{Increasing per-frame token budget leads to substantial declines in average layer-wise recall across a variety of different questions.}
    \label{fig:retrieval_analysis_recall}
\end{minipage}
\hfill
\begin{minipage}{0.48\linewidth}
    \centering
    \includegraphics[width=\linewidth]{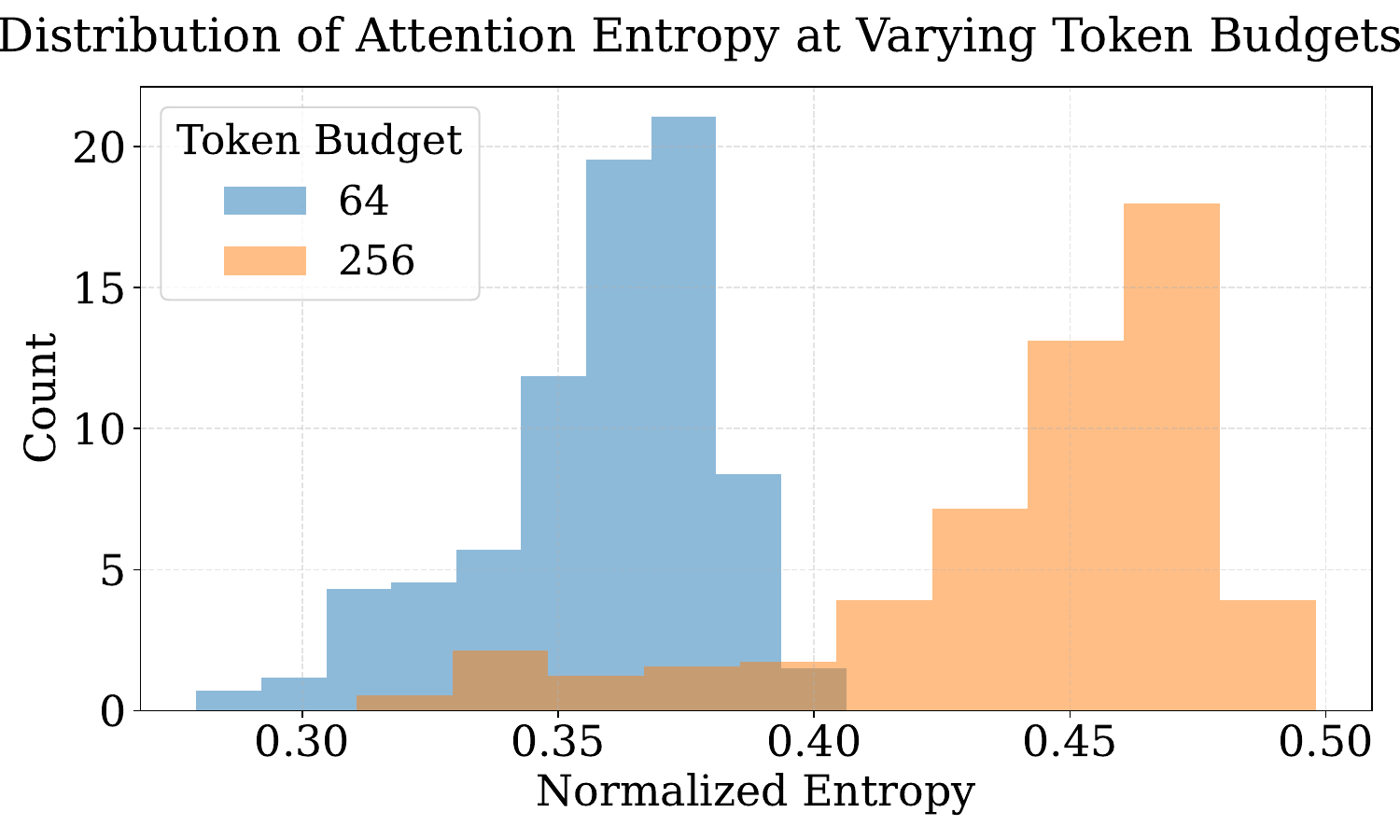}
    \caption{Normalized entropy for 677 sliding windows under 2 separate token budgets. At the higher token budget, sliding window attention tends to exhibit higher entropy, suggesting a struggle to focus on relevant frames.}
    \label{fig:attention_scores}
\end{minipage}
\end{figure*}

\paragraph{Retrieval quality decreases as we increase the token budget.}
In Figure~\ref{fig:retrieval_analysis_recall}, we illustrate how increasing per-frame token budget impacts average layer-wise recall performance. We observe a declining trend, where increasing tokens per frame consistently leads to a drop in average recall.
We confirm this trend across different question types, such as entity and text perception.
Notably, increasing the number of tokens from 64 to 128 leads to a ${\sim}$5\% drop in recall across all question categories.
On average, increasing token budget from 64 to 512 tokens results in a 7\% drop in recall.

\begin{figure*}[t]
\centering

\begin{minipage}[t]{0.54\linewidth}
    \vspace{0pt} %
    \centering
    \includegraphics[width=\linewidth]{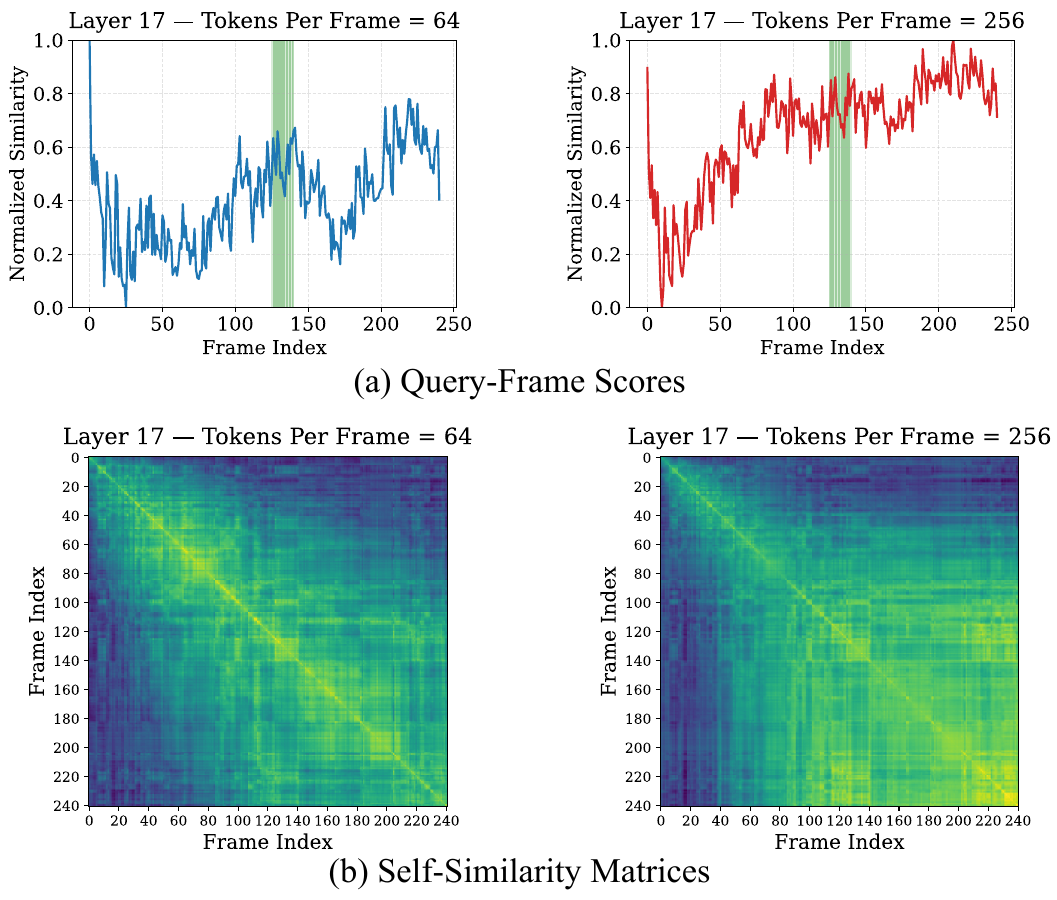}
    \caption{In (a), we identify a systematic trend where query-frame similarity scores progressively increase across the video. In (b), self-similarity maps of the key representations become more redundant as we increase tokens per frame.}
    \label{fig:retrieval_analysis_sims}
\end{minipage}
\hfill
\begin{minipage}[t]{0.42\linewidth}
    \vspace{0pt} %
    \centering

    \includegraphics[width=\linewidth]{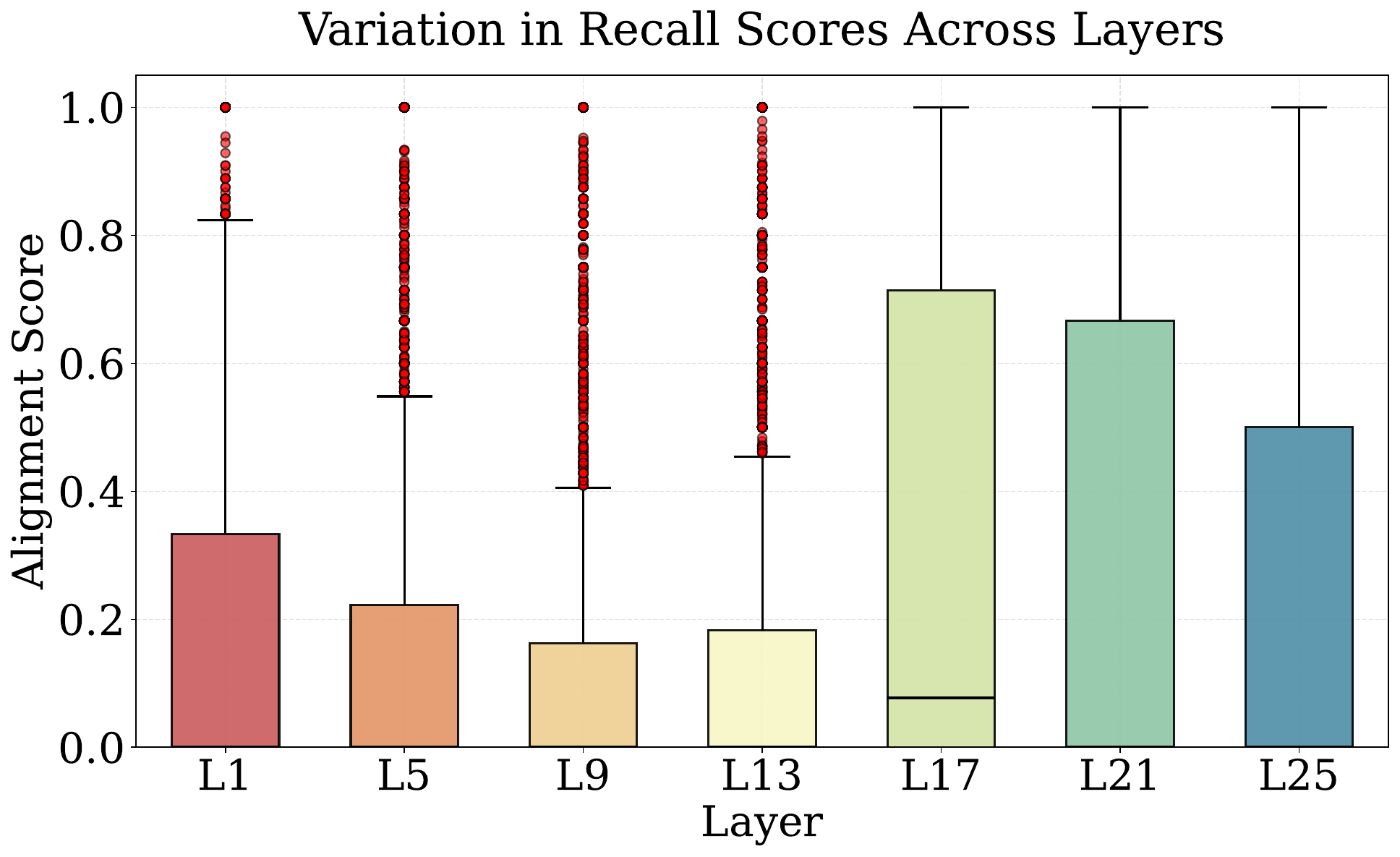}
    \captionof{figure}{Recall for retrieving CG-Bench ``clue'' frames shows high variance across layers and is generally low.}
    \label{fig:layer_retrieval_bad}

    \vspace{0.8em}

    \resizebox{\linewidth}{!}{
    \begin{tabular}{@{}l ccc@{}}
    \toprule
    Method & Num Tokens & CG-Bench & LVBench \\
    \midrule
    Qwen2.5-VL-7B   & 64 & 34.27 & \textbf{37.25}\\
    + Unif. Retrieval ($0.5$ FPS $\rightarrow 128$) & 64 & \textbf{35.23} & 37.06\\
    \midrule
    Qwen2.5-VL-7B   & 128 & \textbf{37.02} & \textbf{39.19} \\
    + Unif. Retrieval ($0.5$ FPS $\rightarrow 128$) & 128 & 37.00 & 36.60 \\
    \midrule
    Qwen2.5-VL-7B   & 256 & \textbf{38.43} & \textbf{41.51}\\
    + Unif. Retrieval ($0.5$ FPS $\rightarrow 128$) & 256 & 37.45 & 37.96\\
    \bottomrule
    \end{tabular}
    }

    \captionof{table}{QA performance with KV-cache memory (uniform-retrieval) degrades at higher token budgets. ($0.5$ FPS $\rightarrow 128$) denotes sampling 128 frames at 0.5 FPS.}
    \label{tab:rekv_problems}

\end{minipage}

\end{figure*}

\paragraph{Retrieval at higher token budgets fails due to temporal bias.}
We plot the similarity scores for frames encoded at 64 tokens per frame and at 256 tokens per frame, labeling the ground-truth segment in green (Figure~\ref{fig:retrieval_analysis_sims}, top).
For frames encoded at lower token budget, the similarity score peaks at the ground-truth location, indicating successful retrieval.
However, at 256 tokens per frame, the query-frame similarity progressively increases across the video, biasing frame selection toward the end of the video.

\paragraph{Self-similarity matrices confirm existence of temporal bias.}
Since query--frame retrieval depends critically on the quality of the key representations stored in the KV-cache, we analyze the self-similarity of representative frame vectors under different per-frame token budgets.
We compute frame-wise self-similarity matrices for 64 and 256 tokens per frame, and visualize representative examples in Figure~\ref{fig:retrieval_analysis_sims} (bottom).
Strikingly, at higher token budgets, representative vectors from different frames become substantially more similar to one another, indicating increased redundancy and reduced discriminability.

\paragraph{Sliding window attention is less selective at higher token budgets.}
We hypothesize that at higher token budgets, the attention mechanism is unable to distinguish relevant spatiotemporal information from each frame.
To validate this, we measure the normalized entropy of the attention scores in the sliding window during video encoding for both a token budget of 64 and 256.
The distribution over 677 sliding windows (Figure~\ref{fig:attention_scores}) confirms that at higher token budget, sliding window attention exhibits greater entropy, indicating less selective behavior.
This leads to degraded feature retrievals, which cascades into failures in question answering.

\paragraph{Increasing per-frame token budget deteriorates question-answering capabilities.}
As shown in Table~\ref{tab:rekv_problems}, at 64 tokens per frame, uniform-retrieval on CG-Bench exceeds the uniform sampling setting by +0.95\% and is roughly equal on LVBench.
However, increasing token budgets leads to significant degradation with a 2.59\% drop at 128 tokens and a 3.55\% drop at 256 tokens on LVBench. 
This highlights that encoding the same frame representations at higher token budgets harms performance.

\subsection{Layer-wise Internal Retrieval is Unreliable}
\label{subsec:unreliable}

ReKV introduces both internal and external mechanisms for retrieving relevant video information from the KV-cache.
Internal retrieval relies on the pre-trained MLLM's internal attention maps for measuring the relevance of each stored frame to the question.
Alternatively, external retrieval uses a pre-trained vision-language model such as CLIP to generate query-frame scores.
ReKV contends that internal retrieval is more robust, with each layer acting as an independent expert.
To isolate from the effect of larger token budgets, all analysis here uses 64 tokens per frame.
We assess each layer's effectiveness using its recall score, shown for a selection of layers in Figure~\ref{fig:layer_retrieval_bad}.
We observe two notable trends: recall performance drastically changes depending on the layer, and it is highly variable even within a single layer.
Some layers have a median recall of 0, indicating that the majority of the time they are unable to retrieve any relevant frames.
This underscores the need for a more stable retrieval mechanism that can complement layer-wise information.

\section{Approach}
\label{sec:Approach}
\begin{figure*}[t]
\begin{center}
\includegraphics[width=\linewidth]{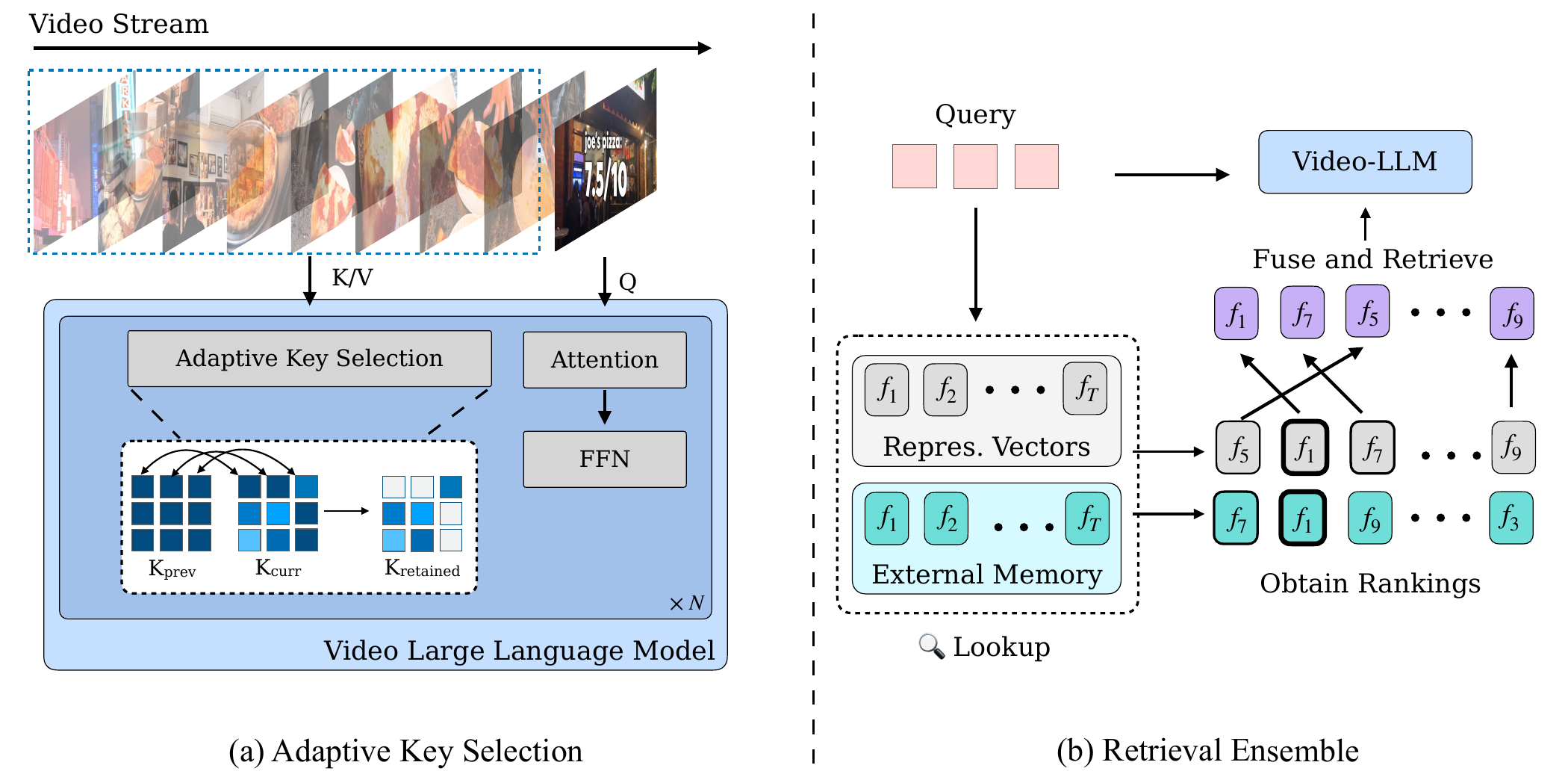}
\caption{(a) Our model processes the actual video stream a single time only to encode video features. We perform adaptive key selection to construct a sparse KV-cache. (b) MemStream uses matching with the question tokens to retrieve the relevant K/V features from the cache. Since we select features on a frame-wise basis, we can perform this retrieval with arbitrary vision-language models. We leverage this to construct a training-free retrieval ensemble for the retrieval.}
\label{fig:method}
\end{center}
\end{figure*}

We introduce MemStream, a training-free unified framework for scalable processing and retrieval of dense video streams.
As shown in Figure~\ref{fig:method}, our approach is split into two stages: encoding and retrieval.
During encoding, we replace dense sliding window attention with Adaptive Key Selection (AKS), a sparse compression and selection strategy to identify and preserve critical video information in the sliding window.
This has a twofold benefit: it preserves local spatiotemporal information across frames, improving KV-cache quality for both retrieval and question-answering, and it reduces latency and memory usage.
When retrieving from the KV-cache memory, we propose a retrieval ensemble strategy that leverages external models to aid and improve internal retrieval quality.

\subsection{Adaptive Key Selection for Sparse Sliding-Window Attention}
\label{subsec:adaptive}
Building on our analysis in Section~\ref{subsec:scaling}, we propose an adaptive compression and selection strategy for sparse sliding-window attention.
We emphasize that this selection occurs only for the attention process, while the full key features are stored and processed in the KV-cache.
Recall that the sliding window at layer $i$ is defined as
$W_t^i = \{(K_j^i, V_j^i)\}_{j=t-\omega}^{t-1}$.
Our goal is to compress and select a representative subset $W_{\text{comp}}^i \subset W_t^i$ that captures its critical spatiotemporal information.
Adaptive Key Selection (AKS) identifies and eliminates temporal redundancy in the sliding window.
Concretely, for each pair of adjacent key features $K^i_t$ and $K^i_{t-1}$, we wish to only retain the patches that are most unique to $K^i_t$.
To this end, we compute patch-wise cosine similarity between corresponding spatial tokens and select the top-$k$ least similar (i.e., most distinctive) patch features from $K_t^i$, where $k$ is fixed. 
We note that this selection process occurs before applying positional embeddings, ensuring that the redundancy calculation is only done based on shared semantics of the patch features. 

\subsection{KV-Cache Retrieval via Retrieval Ensemble}

While ReKV explores external retrievers such as CLIP for query--frame retrieval, it ultimately discards these in favor of purely internal retrieval.
We contend that internal and external retrieval strategies are complementary and can mutually reinforce one another.
Furthermore, external signals can help stabilize internal signals for more consistent layer-wise performance.

In particular, intermediate KV features encode rich contextual information accumulated across the video stream, but may lack fine-grained spatial or motion-specific cues.
Conversely, frame- or clip-level features extracted from vision--language encoders often capture salient semantic details, yet lack access to broader temporal context.

To this end, we propose a training-free retrieval ensemble design that fuses internal attention-based signals with external vision model retrieval using reciprocal rank fusion (RRF).
This approach allows strong retrieval signals from one expert to compensate for weaker signals from another, while incorporating complementary information from both internal and external representations.

In the internal retrieval strategy, layer-wise query-frame scores $S_{\text{internal}}^i \in \mathbb{R}^{T}$ are computed between the question embedding at layer $i$ and the stored representative frame vectors $\{\mathbf{k_j^i}\}_{j=1}^T$.
We follow a similar procedure when using an external vision-language model $E$.
We denote $E_{\text{vis}}$ as the visual encoder and $E_{\text{text}}$ as the corresponding text encoder.
While encoding, we compute a frame-wise feature $x_t \in \mathbb{R}^{1 \times d} = E_{\text{vis}}(f_t)$ for each frame $f_t$ in the video stream.
During question-answering, we compute question $Q$'s embedding $q = E_{\text{text}}(Q)$ and the cosine-similarity between $q$ and all frame embeddings to obtain $S_{\text{external}} \in \mathbb{R}^{T}$.

Rather than fusing raw scores, we take inspiration from information retrieval literature and utilize reciprocal-rank fusion (RRF)~\cite{cormack2009reciprocal}.
Formally, let
\[
R^i = \{ R_{\text{internal}}^i,\, R_{\text{external}} \}
\]
denote the set of retrieval rankings at layer $i$, where
$R_{\text{internal}}^i$ is the ranking produced by internal retrieval and
$R_{\text{external}}$ is the ranking produced by external retrieval.
The RRF score for frame $f_t$ at layer $i$ is defined as
\[
\mathrm{RRFScore}^i(t)
= \sum_{r \in R^i} \frac{1}{k + r(t)},
\]
where $r(t)$ denotes the rank assigned to frame $f_t$ by ranking $r$, and $k$ is a fixed constant.
This scoring function reinforces agreement between rankings while preventing outliers from having too large an effect.
We select the top-$k$ KV features from the cache at each layer using the computed RRF scores.

\begin{table*}[t]
\centering
\begin{minipage}{\linewidth}
\centering
\caption{\textbf{Offline VQA Results.} Effectiveness of MemStream for long video question answering. We highlight the best performance in \textbf{bold}. For VideoMME, we use the ``Long'' subset only. $\dagger$ indicates reproduced results using the available codebase.}
\label{tab:main_results}
\resizebox{\linewidth}{!}{
\begin{tabular}{>{\kern-\tabcolsep}l cc cc c ccc<{\kern-\tabcolsep}}
\toprule
\multicolumn{6}{c}{Settings} & \multicolumn{3}{c}{Benchmarks} \\
\cmidrule(lr){1-6} \cmidrule(lr){7-9}
Model & $N_{\text{tokens}}$ & $N_{\text{frames}}$ & Encoding & Retrieval & Train Free? & CG-Bench & LVBench & VideoMME \\
\midrule
\multicolumn{9}{c}{\textit{Offline Video Question Answering}} \\
\midrule
InternVL3-8B & & & & & & & & \\
\quad + Uniform Sample & 8K & 32 & N/A & N/A & \Checkmark & 38.60 & -- & -- \\
Qwen2.5-VL-7B & & & & & & & & \\
\quad + Uniform Sample & 17K & 128 & N/A & N/A & \Checkmark & 38.43 & 41.51 & 55.67 \\
\quad + Uniform Sample & 68K & 512 & N/A & N/A & \Checkmark & 41.48 & 40.03 & 52.11 \\
\midrule
\multicolumn{9}{c}{\textit{Online Video Question Answering}} \\
\midrule
LLaVA-OneVision & & & & & & & & \\
\quad + rLiVS~\cite{dorovatas2025recurrent} & 10K & ($\small\text{0.5 FPS} \to 51$) & Full & Internal & \Checkmark & 33.10 & -- & -- \\
\quad + ReKV~\cite{di2025rekv} & 15K & ($\small\text{0.5 FPS} \to 64$) & Full & Internal & \Checkmark & 33.90 & 39.90 & \textbf{47.22} \\
\rowcolor{blue!10}
\quad + MemStream (Ours) & 15K & ($\small \text{0.5 FPS} \to 64$) & AKS & Ensemble & \Checkmark & \textbf{35.93} & \textbf{44.93} & 47.00 \\
Qwen2-VL-7B & & & & & & & & \\
\quad + Flash V-Stream~\cite{zhang2025flash} & 12K & ($\small \text{1 FPS} \to \leq \text{12K tokens}$) & N/A & N/A & \XSolidBrush & -- & 42.00 & -- \\
VideoXL-2 & & & & & & & & \\
\quad + TimeScope~\cite{liu2025timescope} & -- & ($\small \text{1 FPS} \to \leq \text{800}$) & -- & -- & \XSolidBrush & 38.47 & -- & -- \\
Qwen2.5-VL-3B & & & & & & & & \\
\quad + ReKV~\cite{di2025rekv} & 17K & ($\small \text{0.5 FPS} \to 128$) & Full & Internal & \Checkmark & 32.59 & 38.54 & 47.78 \\
\rowcolor{blue!10}
\quad + MemStream (Ours) & 17K & ($\small \text{0.5 FPS} \to 128$) & AKS & Ensemble & \Checkmark & \textbf{39.31} & \textbf{46.16} & \textbf{50.44} \\
InternVL3-8B & & & & & & & & \\
\quad + ReKV~\cite{di2025rekv} & 17K & ($\small \text{0.5 FPS} \to 128$) & Full & Internal & \Checkmark & 41.71 & 43.31 & 54.00 \\
\rowcolor{blue!10}
\quad + MemStream (Ours) & 17K & ($\small \text{0.5 FPS} \to 128$) & AKS & Ensemble & \Checkmark & \textbf{44.75} & \textbf{49.19} & 54.00 \\
Qwen2.5-VL-7B & & & & & & & & \\
\quad + TimeChat-Online~\cite{yao2025timechat} & -- & ($\small \text{1 FPS} \to \text{54\% tokens}$) & N/A & N/A & \XSolidBrush & -- & -- & 52.40 \\
\quad + StreamAgent~\cite{yang2025streamagent} & 32 TPF & ($\small \text{1 FPS}$) & Full & Internal & \Checkmark & -- & -- & 50.60 \\
\quad + FluxMem~\cite{xie2026fluxmem} & 64 TPF & ($\small \text{1 FPS}$) & N/A & N/A & \Checkmark & 38.14$^{\dagger}$ & 44.22$^{\dagger}$ & 54.00 \\
\quad + Hermes~\cite{zhang2026hermeskvcachehierarchical} & 6K & ($\small \text{1 FPS}$) & N/A & N/A & \Checkmark & 39.64$^{\dagger}$ & 38.86$^{\dagger}$ & \textbf{54.44} \\
\quad + ReKV~\cite{di2025rekv} & 17K & ($\small \text{0.5 FPS} \to 128$) & Full & Internal & \Checkmark & 36.17 & 39.64 & 51.78 \\
\rowcolor{blue!10}
\quad + MemStream (Ours) & 17K & ($\small \text{0.5 FPS} \to 128$) & AKS & Ensemble & \Checkmark & \textbf{44.19} & \textbf{48.10} & 54.22 \\
Qwen3-VL-8B & & & & & & & & \\
\quad + ReKV~\cite{di2025rekv} & 39K & ($\small \text{0.5 FPS} \to 128$) & Full & Internal & \Checkmark & 52.84 & 51.45 & \textbf{59.44} \\
\rowcolor{blue!10}
\quad + MemStream (Ours) & 39K & ($\small \text{0.5 FPS} \to 128$) & AKS & Ensemble & \Checkmark & \textbf{54.78} & \textbf{55.65} & 59.11 \\
\bottomrule
\end{tabular}
}
\end{minipage}

\vspace{1em} %

\begin{minipage}{\linewidth}
\centering
\caption{\textbf{StreamingVQA Results.} Evaluation of MemStream on RVS-Ego and RVS-Movie, measuring answer accuracy and quality, and runtime and memory usage. Following ReKV, we use GPT-3.5-Turbo as our LLM judge.}
\label{tab:streamingvqa_results}
\resizebox{\linewidth}{!}{
\begin{tabular}{>{\kern-\tabcolsep}lcccccccccc<{\kern-\tabcolsep}}
\toprule
\multicolumn{3}{c}{Settings}
& \multicolumn{2}{c}{RVS-Ego}
& \multicolumn{2}{c}{RVS-Movie}
& \multicolumn{2}{c}{Running Speed}
& \multicolumn{2}{c}{Memory Usage} \\
\cmidrule(lr){1-3}
\cmidrule(lr){4-5}
\cmidrule(lr){6-7}
\cmidrule(lr){8-9}
\cmidrule(lr){10-11}
Model & Encoding & Retrieval
& Acc & Score & Acc & Score
& Video Enc. & Latency
& GPU & KV-Cache \\
\midrule
Qwen2.5-VL-7B & & & & & & & & & & \\
\quad ReKV~\cite{di2025rekv}
& Full & Internal
& 64.2 & 4.00 & \textbf{61.6} & \textbf{3.65}
& 15.10 FPS & 1.7s & 29 GB & 11.1 GB/h \\
\rowcolor{blue!10}
\quad MemStream (Ours)
& AKS & Internal
& \textbf{67.8} & \textbf{4.01} & 59.1 & 3.60
& 12.44 FPS & 1.6s & 29 GB & 11.1 GB/h \\
\rowcolor{blue!10}
\quad MemStream (Ours)
& AKS & Ensemble
& 67.4 & \textbf{4.01} & 59.7 & 3.60
& 10.44 FPS & 1.6s & 32 GB & 11.1 GB/h \\
\bottomrule
\end{tabular}
}
\end{minipage}
\end{table*}

\section{Experiments}
\label{sec:experiments}

\subsection{Benchmark Datasets}
We evaluate on a number of long-form video understanding benchmarks, including both offline and online benchmarks. Details are in Table~\ref{tab:datasets} of the Appendix.
For offline evaluation, we use the multiple-choice subset of CG-Bench~\cite{chen2024cg}, which evaluates a model's ability to retrieve relevant segments for question-answering, 
LVBench~\cite{wang2025lvbench} for very long video understanding, and the ``Long'' subset of VideoMME~\cite{fu2025video}, containing 300 videos with an average duration of 41 mins. 
For online evaluation, we use RVS-Ego and RVS-Movie~\cite{zhang2024flash}, which evaluate streaming VQA via open-ended question-answering scored with the LLM-as-a-Judge paradigm. 

\subsection{Implementation Details}

We integrate MemStream with LLaVA-OneVision-7B, InternVL3-8B, Qwen2.5-VL-3B/7B, and Qwen3-VL-8B. 
Following ReKV, videos are processed at 0.5 FPS with ${\sim}256$ tokens per frame.
For Qwen2.5-VL, we set a minimum of 200 tokens under dynamic resolution, a 17K-token sliding window (roughly 64--68 frames), and 64 retrieved frame features for question answering. 
For Qwen3-VL-8B, we evaluate with a token budget of approximately 512 tokens per frame, with a minimum of 482 tokens per frame, setting the sliding-window limit to 39K tokens.
For external retrieval, we evaluate CLIP and $\text{PE}_{\text{core}}$~\cite{bolya2025perception} with ViT-L backbones and set the reciprocal rank fusion constant $k$ to 60 following~\cite{cormack2009reciprocal}.

\subsection{Results}

\begin{figure}[t]
\begin{center}
\includegraphics[width=0.9\linewidth]{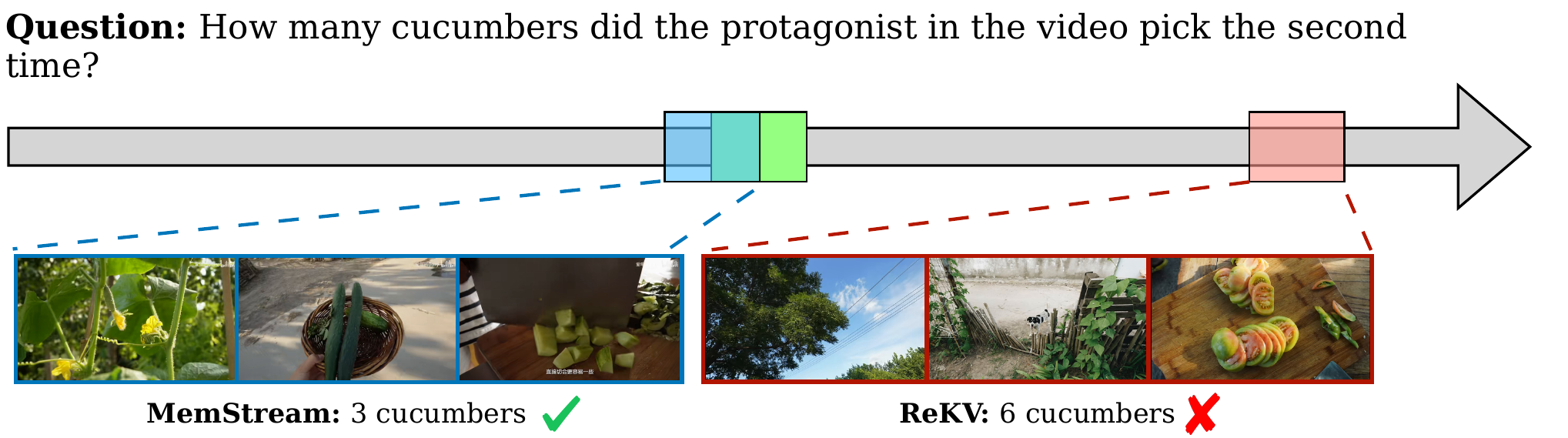}
\caption{\textbf{Qualitative Example.} We compare ReKV's (red) and MemStream's (blue) respective retrievals and answers. The ground-truth segment is shown in green.}
\vspace{-1em}
\label{fig:qualitative_results}
\end{center}
\end{figure}

We show offline results in Table~\ref{tab:main_results} and component-wise effects in Table~\ref{tab:component_breakdown}.
On Qwen2.5-VL, our method significantly outperforms ReKV and recent baselines HERMES~\cite{zhang2026hermeskvcachehierarchical} and FluxMem~\cite{xie2026fluxmem}.
AKS alone improves CG-Bench by 5.5\% and LVBench by 4.1\%.
Adding our retrieval ensemble further improves performance by 2.6\% on CG-Bench and 4.3\% on LVBench.
External retrieval also provides substantial gains, but our combined strategy surpasses it by 2.3\% on both benchmarks.
On VideoMME, our method improves over ReKV by about 2.4\%.
External retrieval alone slightly degrades performance, likely because VideoMME emphasizes holistic understanding, while external retrieval mainly helps find key frames.
We show an example of the improved retrieval using a sample from CG-Bench in Figure~\ref{fig:qualitative_results}.
We next examine online accuracy, quality, latency, and memory usage in Table~\ref{tab:streamingvqa_results}.
MemStream outperforms ReKV on RVS-Ego by 3.2\% with minimal latency change, but drops by 2\% on RVS-Movie, likely due to aggressive compression, as discussed in Section~\ref{subsec:rvs_movie_discussion}.

\subsection{Ablation Study}

\paragraph{Impact of Encoding Strategy} We explore sparse sliding-window attention designs methodically, evaluating static and dynamic selection and compression strategies at both patch- and frame-level granularity. 
For static strategies, we consider average pooling (\textbf{A.1}), which aggregates local spatial information within a $k \times k$ kernel, dilated sampling (\textbf{A.2}), which subsamples spatial tokens by selecting every $k^{\text{th}}$ pixel feature along both height and width dimensions, and uniform frame sampling within the sliding window (\textbf{A.3}). 
For dynamic strategies, in addition to Adaptive Key Selection (\textbf{AKS}), we design three further compression variants: token-merging (\textbf{B.1}), inspired by ToME~\cite{bolya2023tokenmergingvitfaster}, which identifies and compresses redundant patches within each $K_t^i$ without merging across time; frame-wise $k$-means clustering (\textbf{B.2}), which identifies representative centroid frames; and a temporal redundancy strategy (\textbf{B.3}), which retains only frames exhibiting the greatest change over time, computed as the average cosine similarity between adjacent key features. 
Table~\ref{tab:encoding_ablations} shows results of these static and dynamic encoding strategies, validating our decision.
AKS achieves the best overall results despite having the highest compression rate. 
More ablation studies can be found in Sections~\ref{ablation:topk} and~\ref{subsec:compression_rate}.

\begin{table*}[t]
\begin{minipage}[t]{0.54\linewidth}
    \centering
    \caption{\textbf{Encoding Strategy Ablations.} Comparison of patch-wise and frame-wise encoding strategies across benchmarks.}
    \label{tab:encoding_ablations}
    \resizebox{\linewidth}{!}{
    \begin{tabular}{@{}lllcccc@{}}
    \toprule
    Encoding Type & Strategy & \makecell{Comp.\\Rate} & CG-Bench & LVBench & VideoMME \\
    \midrule
    & \text{Full} & ${\sim}1\times$ & 36.17 & 39.64 & 51.78 \\
    \midrule
    \multirow{4}{*}{Patch-wise}
    & \text{A.1} & ${\sim}4\times$ & 40.02 & 40.41 & 51.78 \\
    & \text{A.2} & ${\sim}16\times$ & 41.15 & 42.93 & 52.56 \\
    & \text{B.1} & ${\sim}12\times$ & \textbf{42.18} & 43.06 & 52.89 \\
    & \text{AKS} & ${\sim}16\times$ & 41.63 & \textbf{43.77} & \textbf{54.56} \\
    \midrule
    \multirow{3}{*}{Frame-wise}
    & \text{A.3} & ${\sim}8\times$ & \textbf{41.63} & 42.35 & \textbf{53.78} \\
    & \text{B.2} & ${\sim}8\times$ & 40.48 & 42.41 & 52.33 \\
    & \text{B.3} & ${\sim}8\times$ & 41.22 & \textbf{43.25} & 52.22 \\
    \bottomrule
    \end{tabular}
    }
\end{minipage}
\hfill
\begin{minipage}[t]{0.44\linewidth}
    \centering
    \caption{\textbf{Breakdown of Performance Gains.} AKS and external retrieval each independently improve performance; combining both with our retrieval ensemble achieves the best results.}
    \label{tab:component_breakdown}
    \resizebox{\linewidth}{!}{
    \begin{tabular}{llccc}
    \toprule
    Encoding & Retrieval & CG-Bench & LVBench & VideoMME \\
    \midrule
    Full & Internal   & 36.17 & 39.64 & 51.78 \\
    AKS  & Internal   & 41.63 & 43.77 & \textbf{54.56} \\
    Full & External   & 41.77 & 45.84 & 50.89 \\
    AKS  & External   & 43.21 & 47.19 & 52.33 \\
    \rowcolor{blue!10}
    AKS  & Ensemble   & \textbf{44.19} & \textbf{48.10} & 54.22 \\
    \bottomrule
    \end{tabular}
    }
\end{minipage}

\vspace{1em}

\begin{minipage}[t]{0.54\linewidth}
    \centering
    \caption{\textbf{Retrieval Strategy Ablation.} Comparison of internal, external, and ensemble retrieval. We use the AKS encoding strategy and RRF fusion, with equal weights for the internal and external rankings.}
    \label{tab:retrieval_strategy_ablation}
    \resizebox{\linewidth}{!}{
    \begin{tabular}{lccccc}
    \toprule
    Retrieval Strategy & External Model & CG-Bench & LVBench & VideoMME\\
    \midrule
    Internal Only & -- & 41.63 & 43.77 & 54.56\\
    External Only & CLIP & 42.39 & 47.45 & 53.67 \\
    External Only & $\text{PE}_{\text{core}}$ & 43.21 & 47.19 & 52.33 \\
    \rowcolor{blue!10}
    Ensemble (Ours) & CLIP & 43.84 & 47.39 & \textbf{55.22} \\
    \rowcolor{blue!10}
    Ensemble (Ours) & $\text{PE}_{\text{core}}$ & \textbf{44.19} & \textbf{48.10} & 54.22 \\
    \bottomrule
    \end{tabular}
    }
\end{minipage}
\hfill
\begin{minipage}[t]{0.42\linewidth}
    \centering
    \caption{\textbf{Fusion Strategy Ablation.} Comparison of early- and late-fusion methods for combining internal and external retrieval signals. We use $\text{PE}_{\text{core}}$ as the external encoder and the AKS encoding strategy.}
    \label{tab:fusion_strategy_ablation}
    \resizebox{\linewidth}{!}{
    \begin{tabular}{lccc}
    \toprule
    Fusion Method & CG-Bench & LVBench & VideoMME \\
    \midrule
    L2-Concat & 43.57 & 48.03 & 52.89 \\
    \rowcolor{blue!10}
    RRF (Ours) & \textbf{44.19} & \textbf{48.10} & \textbf{54.22} \\
    \bottomrule
    \end{tabular}
    }
\end{minipage}
\end{table*}

\paragraph{Impact of Retrieval Strategy} Table~\ref{tab:retrieval_strategy_ablation} shows that using CLIP or $\text{PE}_{\text{core}}$ for retrieval results in significant improvements for both CG-Bench and LVBench.
Optimal results are achieved when combining $\text{PE}_{\text{core}}$ with internal retrieval via our retrieval ensemble.
Table~\ref{tab:fusion_strategy_ablation} shows that our reciprocal rank fusion (RRF) is superior to simple concatenation (L2-Concat) across all three offline benchmarks.

\paragraph{Impact of Token Budget}
Table~\ref{tab:token_scaling} scales Qwen2.5-VL-7B from 256 to 768 tokens per frame. AKS outperforms full encoding at every budget on all three benchmarks, while full encoding degrades as the budget grows. Raising the compression rate to ${\sim}25\times$ at 512 and 768 tokens helps mitigate the increased redundancy at these budgets and recovers most of the loss on CG-Bench and LVBench. 

\paragraph{Impact across Video-LLMs}
Table~\ref{tab:backbone_generalization} provides a breakdown of each component for LLaVA-OV-7B and InternVL3. 
Internal retrieval is the primary bottleneck on these backbones, so AKS sees limited gains. 
When retrieval is stabilized with an external encoder, AKS matches or surpasses full encoding.
We show that AKS improves internal recall for all four backbone families in Table~\ref{tab:backbone_recall}.  

\begin{table*}[t]
\begin{minipage}[t]{0.48\linewidth}
    \centering
    \caption{\textbf{Impact of Scaling Token Budget.} AKS maintains performance over ReKV when scaling tokens per frame to higher budgets. $^{*}$768 tokens per frame is evaluated on videos whose native resolution supports this budget without upsampling.}
    \label{tab:token_scaling}
    \resizebox{\linewidth}{!}{%
    \begin{tabular}{@{}llcccc@{}}
    \toprule
    Tokens/Frame & Encoding & \makecell{Comp.\\Rate} & CG-Bench & LVBench & VideoMME \\
    \midrule
    \multirow{2}{*}{${\sim}256$} & Full & $1\times$ & 36.17 & 39.64 & 51.78 \\
    & AKS & ${\sim}16\times$ & \textbf{41.63} & \textbf{43.77} & \textbf{54.56} \\
    \midrule
    \multirow{3}{*}{${\sim}512$} & Full & $1\times$ & 35.41 & 38.35 & 51.00 \\
    & AKS & ${\sim}16\times$ & 42.06 & 42.16 & 52.67 \\
    & AKS & ${\sim}25\times$ & \textbf{42.13} & \textbf{43.32} & \textbf{53.22} \\
    \midrule
    \multirow{3}{*}{${\sim}768^{*}$} & Full & $1\times$ & 35.55 & 35.91 & 50.06 \\
    & AKS & ${\sim}16\times$ & 41.37 & 39.55 & \textbf{53.56} \\
    & AKS & ${\sim}25\times$ & \textbf{41.76} & \textbf{42.90} & 52.56 \\
    \bottomrule
    \end{tabular}
    }
\end{minipage}
\hfill
\begin{minipage}[t]{0.48\linewidth}
    \centering
    \caption{\textbf{Generalization Across Video-LLMs.} We observe that our gains generally transfer to LLaVA-OV and InternVL3.}
    \label{tab:backbone_generalization}
    \resizebox{\linewidth}{!}{%
    \setlength{\tabcolsep}{7.0pt}%
    \begin{tabular}{@{}ll cc cc@{}}
    \toprule
    & & \multicolumn{2}{c}{LLaVA-OV-7B} & \multicolumn{2}{c}{InternVL3-8B} \\
    \cmidrule(lr){3-4} \cmidrule(lr){5-6}
    Encoding & Retrieval & CG-Bench & LVBench & CG-Bench & LVBench \\
    \midrule
    Full & Internal & 33.90 & 39.90 & 41.71 & 43.31 \\
    AKS  & Internal & 33.98 & 39.12 & 41.45 & 43.38 \\
    Full & External & 35.49 & 43.51 & \textbf{44.76} & 47.84 \\
    AKS  & External & 35.62 & 44.29 & 44.66 & 49.00 \\
    \rowcolor{blue!10}
    AKS  & Ensemble & \textbf{35.93} & \textbf{44.93} & 44.75 & \textbf{49.19} \\
    \bottomrule
    \end{tabular}
    }

    \vspace{0.2em}

    \caption{\textbf{Recall Across Video-LLMs.} AKS improves internal recall across all backbones.}
    \label{tab:backbone_recall}
    \resizebox{\linewidth}{!}{%
    \setlength{\tabcolsep}{5.4pt}%
    \begin{tabular}{@{}l cccc@{}}
    \toprule
    Encoding & LLaVA-OV-7B & InternVL3-8B & Qwen2.5-VL-7B & Qwen3-VL-8B \\
    \midrule
    Full & 15.72 & 12.47 & 16.27 & 37.97 \\
    \rowcolor{blue!10}
    AKS  & \textbf{18.14} & \textbf{13.47} & \textbf{28.91} & \textbf{38.67} \\
    \bottomrule
    \end{tabular}
    }
\end{minipage}
\end{table*}

\section{Limitations and Future Work}
\label{subsec:limitations}
In this work, we identify two key insights: (1) sparse attention enables effective scaling to larger token budgets, and (2) external vision-language models can complement strong Video-LLMs for video understanding. These findings offer several directions for future work. First, our AKS strategy preserves a fixed ratio of patches; future work could make this ratio adaptive, retaining more patches for dynamic scenes and fewer for static ones. Second, training-based sparse methods may be more robust for scaling to larger token budgets. Finally, while we use fixed weights to combine internal and external retrievals, an adaptive mixture-of-experts-style routing could further improve performance.

\section{Conclusion}

\label{sec:conclusion}
We integrate KV-caching with modern MLLMs to enable efficient long video understanding.
Through detailed analysis, we identify that higher token budgets in KV-cache streaming introduce temporal redundancy and retrieval bias.
To address both, we propose Adaptive Key Selection (AKS) for higher-quality encoded features, and a training-free retrieval ensemble leveraging external model features for more precise temporal grounding.
Our method, MemStream, achieves strong performance on both offline and online long video benchmarks.

\section{Acknowledgements}
This work was partially supported by NSF CAREER Award (\#2238769). 
The authors acknowledge UMD's supercomputing resources made available for conducting this research. 
The U.S. Government is authorized to reproduce and distribute reprints for Governmental purposes notwithstanding any copyright annotation thereon. 
The views and conclusions contained herein are those of the authors and should not be interpreted as necessarily representing the official policies or endorsements, either expressed or implied, of NSF or the U.S. Government.
The authors would like to thank our colleagues Anubhav Gupta, Namitha Padmanabhan, and Max Ehrlich for their
valuable conversations and feedback.

\clearpage
{\small
\bibliographystyle{ieeetr}
\bibliography{main}
}

\clearpage
\appendix

{
\centering
\Large
\textbf{Going Down Memory Lane: Scaling Tokens for Video
Stream Understanding with Dynamic KV-Cache
Memory} \\ 
     {\small Supplementary Material} \\
\vspace{2.0em}
} 

\startcontents[supplementary]
\printcontents[supplementary]{l}{1}{\section*{Contents}}

\clearpage
\section{Further Preliminaries}

\subsection{KV-Cache for Video Understanding}

\paragraph{Online Video Understanding.}
This work builds an adaptive memory framework for processing a continuous stream of video.
This differs significantly from the conventional offline setting.
Namely, frames are observed and stored in an incremental manner, preventing the model from using future context to encode current frames.
Furthermore, the length of the stream is unknown, rendering uniform frame sampling strategies moot.

\paragraph{Vision-Language Understanding via KV-Cache.}
Modern large language models rely on key--value caching for improving efficiency during generation.
There are two stages of this pipeline.
First, the model performs prefilling; here, the model processes and stores intermediate key--value features of previous tokens, forming the KV-cache.
Then, during decoding, each new token is generated by attending to the tokens in the stored KV-cache.
MLLMs also exploit this structure for storing visual inputs in the KV-cache for use in downstream tasks such as image captioning or question-answering.

ReKV~\cite{di2025rekv} extends this line of work by introducing a KV-cache-based memory for long-video processing.
It incrementally encodes the video stream using sliding-window attention, offloading key--value features to RAM or disk once the frame has exited the window.
During question-answering, ReKV leverages the internal attention of the LLM to retrieve relevant key--value entries from the cache.

\subsection{KV-Cache Size Calculation}

Assuming FP16 precision, the size of the KV-Cache is calculated as follows,
\[
2 \times L \text{ layers} \times T \text{ frames} \times M \text{ tokens/frame} \times H \text{ heads} \times D \text{ dimension} \times 2 \text{ bytes}.
\]

Qwen2.5-VL-7B has $L = 28$ layers with $H = 4$ heads of dimension $D = 128$.
A 1-hour video processed at 0.5 FPS translates to $T = 900$ frames (using temporal-patch-size of 2).
We set the frame-wise token budget between 200 and 256 tokens, resulting in a KV-cache size between 10.3 GB and 13.2 GB.
Our results in Table~\ref{tab:streamingvqa_results} are within this bound.

\section{Hyperparameters and Settings}

\subsection{Benchmark Datasets}

\begin{table}[ht]
\centering
\caption{\textbf{Dataset statistics.} We detail the number of videos, questions, and average length for each benchmark.}
\label{tab:datasets}
\resizebox{0.6\linewidth}{!}{
\begin{tabular}{@{}lccc@{}}
\toprule
Dataset & \# Videos & Avg. Length & \# QA Pairs \\
\midrule
CG-Bench~\cite{chen2024cg} & 1,219 & 27 min. & 12,129 \\
LVBench~\cite{wang2025lvbench} & 103 & 68 min. & 1,549 \\
VideoMME (Long)~\cite{fu2025video} & 300 & 41 min. & 900 \\
RVS-Ego~\cite{zhang2024flash} & 10 & 60 min. & 1,465 \\
RVS-Movie~\cite{zhang2024flash} & 22 & 30 min. & 1,905 \\
\bottomrule
\end{tabular}
}
\end{table}

\subsection{Integrating KV-Cache Memory with Qwen2.5-VL}
\label{subsec:qwen_integration}

Integrating KV-Cache memory with Qwen2.5-VL requires three main considerations.
First, Qwen2.5-VL applies dynamic resolution tokenization, where the number of tokens can vary for different images, and the original resolution of the image is preserved as best as possible.
Second, rather than using a fixed 1D position encoding such as RoPE, Qwen2.5-VL employs the spatiotemporal-aware M-RoPE position encoding strategy.
Lastly, Qwen2.5-VL uses a temporal patch-size of 2 during tokenization, such that one ``frame-feature'' corresponds to 2 frames.
This means that for a video with $N$ frames, the model stores $N_{kv} = N/2$ frame features.

\subsection{Compression Rates}
\label{subsec:compression}

Throughout the paper, the compression rate $r$ denotes the fraction of sliding-window keys that participate in attention during encoding. Patch-wise strategies, including AKS, retain $N/r$ of the $N$ patch features of each frame (i.e., $k = N/r$ in Section~\ref{sec:Approach}), while frame-wise strategies retain $1/r$ of the frames in the window. Compression only affects the keys and values attended to within the sliding window; the full KV features of every frame are still written to the cache and remain available for retrieval. For dynamic-resolution models, $k$ is fixed while $N$ varies across frames, so the rate is approximate (e.g., ${\sim}16\times$ for Qwen2.5-VL at 200--256 tokens per frame). Table~\ref{tab:compression_settings} lists the settings used for the main results. Unless stated otherwise, ablations in Section~\ref{sec:experiments} and in this appendix use Qwen2.5-VL-7B at ${\sim}16\times$. In the token-scaling study (Table~\ref{tab:token_scaling}) we additionally report ${\sim}25\times$ at 512 and 768 tokens per frame, so that the number of retained keys per frame stays close to the 256-token setting. For Qwen3-VL-8B we use a mild ${\sim}1.33\times$ rate (retaining 75\% of patches), since higher rates reduce accuracy on that backbone (Section~\ref{subsec:qwen3vl}).

\begin{table}[h]
\centering
\caption{\textbf{Compression Settings for Main Results.} Token budget, sliding-window size, AKS compression rate, and the resulting number of retained keys per frame feature for each backbone in Table~\ref{tab:main_results}.}
\label{tab:compression_settings}
\resizebox{0.75\linewidth}{!}{
\begin{tabular}{lcccc}
\toprule
Backbone & Tokens/Frame & Sliding Window & AKS Rate & Retained Keys/Frame \\
\midrule
Qwen2.5-VL-3B & ${\sim}256$ & 17K & ${\sim}16\times$ & ${\sim}16$ \\
Qwen2.5-VL-7B & ${\sim}256$ & 17K & ${\sim}16\times$ & ${\sim}16$ \\
InternVL3-8B & 256 & 17K & $16\times$ & 16 \\
LLaVA-OneVision-7B & 196 & 15K & $4\times$ & 49 \\
Qwen3-VL-8B & ${\sim}512$ & 39K & ${\sim}1.33\times$ & ${\sim}384$ \\
\bottomrule
\end{tabular}
}
\end{table}

\section{Experiments Cont.}

All experimental results are conducted with Qwen2.5-VL-7B with 256 tokens per frame at 0.5 FPS, unless otherwise specified. 

\subsection{More Streaming Results}

As our method focuses more on improving retrieval rather than real-time streaming understanding, we evaluate MemStream on the backward-tracing subset of OVOBench~\cite{niu2025ovo}, which tests a streaming model's ability to recall past events via episodic memory (EPM), action sequence identification (ASI), and hallucination detection (HLD).
\begin{table}[h]
\centering
\caption{\textbf{OVOBench Backward-Tracing Results.} We compare MemStream with ReKV along with other strong baselines. $\dagger$ indicates training-based methods}
\label{tab:ovobench}
\resizebox{0.8\linewidth}{!}{
\begin{tabular}{llcccc}
\toprule
Model & Config & ASI & EPM & HLD & Avg \\
\midrule
VideoLLM-online-8B$^\dagger$~\cite{videollm-online} & -- & 22.2 & 18.8 & 12.2 & 17.7 \\
Flash-VStream-7B$^\dagger$~\cite{zhang2024flash} & -- & 39.1 & 37.2 & 5.9 & 27.4 \\
\midrule
\multirow{2}{*}{Qwen2.5-VL-7B} & ReKV & 64.9 & 52.3 & 23.2 & 46.4 \\
& MemStream & 62.2 & 53.9 & 25.3 & 46.9 \\
\bottomrule
\end{tabular}
}
\end{table}

MemStream outperforms ReKV and other streaming baselines on average, with particular gains on episodic memory and hallucination detection. Together with our results on RVS-Ego (+3\%) and offline benchmarks, these results demonstrate that MemStream’s adaptive encoding and retrieval ensemble strategies consistently improve streaming video understanding across diverse settings.

\subsection{Inference Analysis of Retrieval Ensemble}

We first provide a breakdown of MemStream's runtime, explicitly identifying the cost of the external encoder. 
Table~\ref{tab:latency_breakdown} shows that the external encoder accounts for 9.0\% of total runtime (15.9\% of encoding time) and adds about 6 GB of GPU memory, while external retrieval itself is negligible at 0.4\%. Decoding and the MLLM visual backbone dominate the runtime.

\begin{table}[h]
\centering
\caption{\textbf{Latency Breakdown with the External Encoder.} Wall-clock time per stage for Qwen2.5-VL-7B + MemStream (AKS encoding and $\text{PE}_{\text{core}}$ ensemble) on one 60-minute RVS-Ego video with 100 questions at 1 FPS on a single H200 GPU.}
\label{tab:latency_breakdown}
\resizebox{0.6\linewidth}{!}{
\begin{tabular}{llcc}
\toprule
\# & Stage & Time (s) & \% of Runtime \\
\midrule
1 & MLLM visual backbone & 101.6 & 27.5 \\
2 & External visual encoder ($\text{PE}_{\text{core}}$) & 33.2 & 9.0 \\
2b & \quad External retrieval & 1.3 & 0.4 \\
3 & KV-cache encoding & 73.9 & 20.0 \\
4 & Retrieval & 14.2 & 3.8 \\
5 & Decoding & 145.3 & 39.3 \\
\midrule
 & Total & 369.6 & 100 \\
\bottomrule
\end{tabular}
}
\end{table}

In the main paper, we showed that our plug-and-play design can be used to augment smaller VideoLLMs with strong external VL encoders. We investigate this further here.  

\begin{table}[h]
\centering
\caption{\textbf{MemStream on Smaller Models.} Qwen2.5-VL-3B + MemStream outperforms the larger Qwen2.5-VL-7B + ReKV on CG-Bench and LVBench.}
\label{tab:small_model_acc}
\resizebox{0.6\linewidth}{!}{
\begin{tabular}{llccc}
\toprule
Model & Config & CG-Bench & LVBench & VideoMME \\
\midrule
\multirow{2}{*}{Qwen2.5-VL-3B} & ReKV & 32.59 & 38.54 & 47.78 \\
& MemStream & \textbf{39.31} & \textbf{46.16} & \textbf{50.44} \\
\midrule
\multirow{2}{*}{Qwen2.5-VL-7B} & ReKV & 36.17 & 39.64 & 51.78 \\
& MemStream & \textbf{44.19} & \textbf{48.10} & \textbf{54.22} \\
\bottomrule
\end{tabular}
}
\end{table}

\begin{table}[h]
\centering
\caption{\textbf{Efficiency Comparison Across Model Sizes.} All configurations use AKS encoding.}
\label{tab:small_model_efficiency}
\resizebox{0.7\linewidth}{!}{
\begin{tabular}{llcccc}
\toprule
Model & Retrieval & Video Enc. & Latency & GPU & KV-Cache \\
\midrule
Qwen2.5-VL-7B & Internal & 12.44 FPS & 1.6s & 29 GB & 11.1 GB/h \\
Qwen2.5-VL-3B & Ensemble & 10.58 FPS & 0.7s & 21 GB & 7.1 GB/h \\
Qwen2.5-VL-7B & Ensemble & 10.44 FPS & 1.6s & 32 GB & 11.1 GB/h \\
\bottomrule
\end{tabular}
}
\end{table}

Qwen2.5-VL-3B + MemStream outperforms the larger Qwen2.5-VL-7B + ReKV baseline on CG-Bench (39.31 vs 36.17) and LVBench (46.16 vs 39.64). This suggests that a smaller model equipped with MemStream can be a viable alternative to a larger model on resource-constrained devices.

\subsection{RVS-Movie Discussion}
\label{subsec:rvs_movie_discussion}

\begin{table}[h]
\begin{minipage}{0.48\linewidth}
\centering
\caption{\textbf{Effect of AKS Compression Rate on RVS-Movie.}}
\label{tab:aks_compression_rvs}
\resizebox{0.75\linewidth}{!}{
\begin{tabular}{ccc}
\toprule
Comp. Rate & Acc & Score \\
\midrule
${\sim}4\times$ & 61.1 & 3.64 \\
${\sim}16\times$ & 59.1 & 3.60 \\
\bottomrule
\end{tabular}
}
\end{minipage}
\hfill
\begin{minipage}{0.48\linewidth}
\centering
\caption{\textbf{AKS on Low-Dynamic Content (TV \& Music Subset of CG-Bench).}}
\label{tab:aks_tv_music}
\resizebox{0.75\linewidth}{!}{
\begin{tabular}{cc}
\toprule
Encoding & TV \& Music \\
\midrule
Full & 32.73 \\
AKS & 38.77 \\
\bottomrule
\end{tabular}
}
\end{minipage}
\end{table}

In the main paper, we observed that RVS-Movie suffered a drop in performance at a $16 \times$ compression ratio. We hypothesize that this may be due to AKS mistakenly discarding important semantics at high compression rates. To this end, we investigate model performance at a lower compression ratio on RVS-Movie (as shown in Table~\ref{tab:aks_compression_rvs}) and find that reducing the compression ratio from 1/16 to 1/4 boosts accuracy by 2\%. Furthermore, we examine how AKS performs on content with subtle changes via a 1,067 video subset of CG-Bench containing TV and Music content, which is rich in dialogue and facial expressions. Examining Table~\ref{tab:aks_tv_music}, we find that AKS achieves strong gains over the ReKV baseline, suggesting it does not systematically discard important semantic content in low-dynamic scenes. 

\begin{table}[h]
\centering
\caption{{\textbf{Cache vs.\ Retrieval on RVS-Movie.} Accuracy for all four combinations of which KV-cache answers the question (rows) and which retrieval ranking selects the frames (columns), using Qwen2.5-VL-7B at ${\sim}16\times$ compression.}}
\label{tab:rvs_movie_2x2}
\resizebox{0.45\linewidth}{!}{
\begin{tabular}{lcc}
\toprule
 & ReKV Retrievals & AKS Retrievals \\
\midrule
ReKV Cache & 61.6 & \textbf{63.4} \\
AKS Cache & 58.0 & 59.1 \\
\bottomrule
\end{tabular}
}
\end{table}

We disentangle whether AKS degrades the retrievals or the KV-cache representations used for QA by evaluating all four combinations of cache (which representations answer) $\times$ retrievals (which rankings select frames).
Table~\ref{tab:rvs_movie_2x2} evaluates all four combinations of cache and retrieval. 
AKS retrievals improve accuracy for both cache representations, while the AKS cache is consistently lower than the ReKV cache, which localizes the degradation to the encoded representations. 
Failures concentrate on temporal-order questions; we hypothesize that discarding adjacent redundant patches at $16\times$ removes the subtle frame-to-frame changes that establish event order. 
For example, when asked what happens before two ships exchange fire, the ReKV cache correctly describes the crew's preparations, whereas the AKS cache misorders the events and drifts into generic scene description.

\subsection{Impact of Top-$k$ and Sliding-Window Size}
\label{ablation:topk}

In Table~\ref{tab:retrieval_size}, we provide ablations on the impact of the number of retrieved frames on model performance and compare ReKV (Full) and our strategy (AKS). We make two key observations. First, AKS achieves consistent performance gains across all top-k settings. Second, increasing the number of frames for retrieval, generally improves performance. We also ablate the impact of sliding window size on model performance in Table~\ref{tab:sliding_window}, observing that increasing sliding-window size generally boosts performance until a certain point, at which accuracy drops slightly. This could suggest increased redundancy, requiring greater compression.

\begin{table}[h]
\begin{minipage}{0.55\linewidth}
\centering
\caption{\textbf{Effect of Retrieval Size.}}
\label{tab:retrieval_size}
\resizebox{0.8\linewidth}{!}{
\begin{tabular}{llccc}
\toprule
Encoding & Top-$k$ & CG-Bench & LVBench \\
\midrule
\multirow{3}{*}{Full} & 32 & 33.06 & 32.80 \\
& 64 & 34.54 & 34.80 \\
& 128 & 36.17 & 39.64 \\
\midrule
\multirow{3}{*}{AKS} & 32 & 35.05 & 38.02 \\
& 64 & 38.43 & 39.64 \\
& 128 & 41.63 & 43.77 \\
\bottomrule
\end{tabular}
}
\end{minipage}
\hfill
\begin{minipage}{0.38\linewidth}
\centering
\caption{\textbf{Effect of Sliding-Window Size.}}
\label{tab:sliding_window}
\resizebox{\linewidth}{!}{
\begin{tabular}{ccccc}
\toprule
& 8500 & 13000 & 17000 & 25000 \\
\midrule
LVBench & 40.34 & 40.80 & 43.77 & 43.25 \\
\bottomrule
\end{tabular}
}
\end{minipage}
\end{table}

\subsection{Impact of Compression Rate}
\label{subsec:compression_rate}

In Table~\ref{tab:compression_encoding_acc}, we evaluate static selection patterns at compression rates $r \in \{4, 8, 16\}$.
Patch-wise strategies retain $1/r$ tokens per frame, whereas frame-wise strategies retain $1/r$ frames from the sliding window.
Sliding-window compression improves performance over full attention (up to +5.7\% on CG-Bench and +3.6\% on LVBench), with gains peaking at the highest compression factor. Additionally, in Table~\ref{tab:aks_compression}, we evaluate AKS at different compression ratios, observing that increasing the compression ratio generally improves performance. 

\begin{table}[h]
\centering
\caption{\textbf{Effect of Sliding-Window Compression Rate.} We evaluate performance of static attention patterns at varying sliding-window compression rates.}
\label{tab:compression_encoding_acc}
\resizebox{0.6\linewidth}{!}{
\begin{tabular}{llccc}
\toprule
Encoding & \makecell[l]{Comp. \\ Rate} & CG-Bench & LVBench & VideoMME \\
\midrule
\text{Full} & $1\times$ & 36.17 & 39.64 & 51.78 \\
\midrule
\multirow{3}{*}{Pooled (A.1)}
& ${\sim}4\times$ & 40.02 & \textbf{40.41} & \textbf{51.78} \\
& ${\sim}8\times$ & 39.81 & 40.35 & 50.33 \\
& ${\sim}16\times$ & 39.54 & 39.57 & 50.22 \\
\midrule
\multirow{3}{*}{Dilated (A.2)}
& ${\sim}4\times$ & 40.17 & 41.12 & 51.89 \\
& ${\sim}8\times$ & \textbf{41.69} & 42.67 & 51.78 \\
& ${\sim}16\times$ & 41.15 & \textbf{42.93} & \textbf{52.56} \\
\midrule
\multirow{3}{*}{\makecell[l]{\text{Uniform} \\ \text{Sample (A.3)}}}
& ${\sim}4\times$ & 40.88 & 40.87 & 51.78 \\
& ${\sim}8\times$ & 41.63 & 42.35 & \textbf{53.78} \\
& ${\sim}16\times$ & \textbf{41.83} & \textbf{43.19} & 52.89 \\
\bottomrule
\end{tabular}
}
\end{table}

\begin{table}[h!]
\centering
\caption{\textbf{Effect of AKS Compression Rate.} We evaluate performance of AKS at varying compression ratios.}
\label{tab:aks_compression}
\resizebox{0.4\linewidth}{!}{
\begin{tabular}{ccc}
\toprule
Comp. Rate & CG-Bench & LVBench \\
\midrule
${\sim}4\times$ & 40.66 & 43.12 \\
${\sim}9\times$ & 40.47 & 43.38 \\
${\sim}16\times$ & 41.63 & 43.77 \\
\bottomrule
\end{tabular}
}
\end{table}

\subsection{Impact on Recall}

In Table~\ref{tab:cgbench_recall},  we show how AKS and our external retrieval impact recall for different types of questions on CG-Bench. 

\begin{table*}[h!]
\centering
\caption{\textbf{CG-Bench Recall.} We show how frequently the top 64 retrievals overlap the features from the ground truth clue frames in CG-Bench.}
\label{tab:cgbench_recall}
\resizebox{1.0\linewidth}{!}{
\begin{tabular}{@{}llll cccccc@{}}
\toprule
\multicolumn{4}{c}{Settings} &
\multicolumn{6}{c}{CG-Bench Average Recall@64 Scores} \\
\cmidrule{1-4}
\cmidrule(l){5-10}
Model & Method & Encoding & Retrieval & Entity Cognition & Entity Perception & Event Cognition & Event Perception & Text Perception & Mean \\
\midrule
\multicolumn{10}{@{}l}{\textit{Internal Retrieval}} \\
\midrule
& Uniform & Uniform & Internal & 20.57 & 22.20 & 21.98 & 22.29 & 26.12 & 22.24 \\
& ReKV & Full & Internal & 14.49 & 15.59 & 16.03 & 15.22 & 19.63 & 16.27 \\
& MemStream & A.2 ($\times 4$) & Internal & 19.97 & 21.42 & 20.86 & 20.23 & 27.31 & 21.59 \\
& MemStream & A.2 ($\times 16$) & Internal & 24.89 & 27.17 & 26.08 & 26.94 & 35.31 & 27.41 \\
& MemStream & A.3 ($\times 4$) & Internal & 21.83 & 23.64 & 22.38 & 21.84 & 29.95 & 23.36 \\
& MemStream & A.3 ($\times 16$) & Internal & \textbf{26.85} & \textbf{29.51} & 27.24 & 27.57 & \textbf{38.32} & 28.79 \\
\rowcolor{blue!10}
& MemStream & AKS ($\times 4$) & Internal & 20.86 & 22.25 & 21.64 & 21.26 & 29.10 & 22.61 \\
\rowcolor{blue!10}
\multirow{-8}{*}{Qwen2.5-VL-7B} & MemStream & AKS ($\times 16$) & Internal & 26.50 & 28.73 & \textbf{27.35} & \textbf{28.44} & 37.20 & \textbf{28.91} \\
\midrule
\multicolumn{10}{@{}l}{\textit{With External Retrieval}} \\
\midrule
$\text{PE}_{\text{core}}$ ViT-L & -- & -- & -- & \textbf{54.54} & \textbf{63.96} & \textbf{52.43} & \textbf{54.51} & 58.41 & \textbf{55.94} \\
Qwen2.5-VL-7B & MemStream & AKS ($\times 16$) & Ensemble & 51.00 & 58.04 & 48.84 & 51.02 & \textbf{58.52} & 52.32 \\
\bottomrule
\end{tabular}
}
\end{table*}

\subsection{Category Breakdown}

In Tables~\ref{tab:cgbench_breakdown} and~\ref{tab:lvbench_breakdown}, we present the breakdown of performance for various question types. 

\begin{table*}[ht]
\centering
\caption{\textbf{CG-Bench Breakdown.} We breakdown CG-Bench performance for a subset of question-types. We \textbf{bold} the best result and \underline{underline} the second-best.}
\label{tab:cgbench_breakdown}
\resizebox{1.0\linewidth}{!}{
\begin{tabular}{@{}llll cccccc@{}}
\toprule
\multicolumn{4}{c}{Settings} &
\multicolumn{6}{c}{CG-Bench Accuracy} \\
\cmidrule{1-4}
\cmidrule(l){5-10}
Model & Method & Encoding & Retrieval & Entity Cognition & Entity Perception & Event Cognition & Event Perception & Text Perception & Mean \\
\midrule
\multicolumn{10}{@{}l}{\textit{Internal Retrieval}} \\
\midrule
& Uniform & Uniform & Internal & 34.03 & 37.06 & 36.76 & 35.35 & 49.52 & 38.43 \\
& ReKV & Full & Internal & 33.12 & 36.35 & 34.73 & 31.36 & 45.72 & 36.17 \\
& MemStream & A.2 ($\times 4$) & Internal & 35.70 & 39.08 & 39.26 & 36.02 & 52.42 & 40.17 \\
& MemStream & A.2 ($\times 16$) & Internal & 36.61 & 40.14 & 39.26 & \textbf{37.99} & 54.36 & 41.15 \\
& MemStream & A.3 ($\times 4$) & Internal & 36.05 & \underline{40.29} & 39.12 & 36.44 & 54.52 & 40.88 \\
& MemStream & A.3 ($\times 16$) & Internal & \textbf{38.56} & 40.14 & \textbf{40.07} & 37.32 & \textbf{56.87} & \textbf{41.83} \\
\rowcolor{blue!10}
& MemStream & AKS ($\times 4$) & Internal & 36.96 & 39.98 & 39.66 & 36.15 & 52.50 & 40.66 \\
\rowcolor{blue!10}
\multirow{-8}{*}{Qwen2.5-VL-7B} & MemStream & AKS ($\times 16$) & Internal & \underline{38.08} & \textbf{40.66} & \underline{39.86} & \underline{37.70} & \underline{55.01} & \underline{41.63} \\
\midrule
\multicolumn{10}{@{}l}{\textit{With External Retrieval}} \\
\midrule
& MemStream & AKS ($\times 16$) & $\text{PE}_{\text{core}}$ Only & 38.77 & 43.25 & 40.74 & 38.12 & 59.45 & 43.21 \\
\multirow{-2}{*}{Qwen2.5-VL-7B} & MemStream & AKS ($\times 16$) & Ensemble & \textbf{40.03} & \textbf{43.32} & \textbf{43.78} & \textbf{40.01} & \textbf{60.18} & \textbf{44.19} \\
\bottomrule
\end{tabular}
}
\end{table*}

\begin{table*}[h!]
\centering
\caption{\textbf{LVBench Breakdown.} We breakdown LVBench performance across all question-types. We \textbf{bold} the best result and \underline{underline} the second-best.}
\label{tab:lvbench_breakdown}
\resizebox{1.0\linewidth}{!}{
\begin{tabular}{@{}llll ccccccc@{}}
\toprule
\multicolumn{4}{c}{Settings} &
\multicolumn{7}{c}{LVBench Accuracy} \\
\cmidrule{1-4}
\cmidrule(l){5-11}
Model & Method & Encoding & Retrieval & Entity Recog. & Event Understand. & Key Info. Retrieval & Reasoning & Summarizing & Temporal Ground. & Mean \\
\midrule
\multicolumn{10}{@{}l}{\textit{Internal Retrieval}} \\
\midrule
& Uniform & Uniform & Internal & 41.65 & 39.72 & 42.96 & 42.29 & \textbf{43.10} & 39.09 & 41.51 \\
& ReKV & Full & Internal & 38.55 & 39.57 & 43.30 & 40.30 & 29.31 & 35.00 & 39.64 \\
& MemStream & A.2 ($\times 4$) & Internal & 40.33 & 41.11 & 44.33 & 41.29 & 31.03 & 36.82 & 41.12 \\
& MemStream & A.2 ($\times 16$) & Internal & 41.80 & \textbf{42.35} & 45.02 & 41.79 & {34.48} & 39.09 & 42.93 \\
& MemStream & A.3 ($\times 4$) & Internal & 40.47 & 39.88 & 45.36 & 43.28 & 27.59 & 36.82 & 40.87 \\
& MemStream & A.3 ($\times 16$) & Internal & \textbf{43.13} & 40.19 & 48.11 & 40.30 & {34.48} & \textbf{40.45} & \underline{43.19} \\
\rowcolor{blue!10}
& MemStream & AKS ($\times 4$) & Internal & 42.10 & 40.65 & \underline{48.45} & \textbf{46.27} & \underline{37.93} & \underline{40.00} & 43.12 \\
\rowcolor{blue!10}
\multirow{-8}{*}{Qwen2.5-VL-7B} & MemStream & AKS ($\times 16$) & Internal & \underline{42.98} & \underline{41.27} & \textbf{50.52} & \underline{42.79} & {34.48} & \textbf{40.45} & \textbf{43.77} \\
\midrule
\multicolumn{10}{@{}l}{\textit{With External Retrieval}} \\
\midrule
& MemStream & AKS ($\times 16$) & $\text{PE}_{\text{core}}$ Only & \textbf{51.26} & 40.49 & \textbf{58.42} & 41.79 & \textbf{36.21} & 40.91 & 47.19 \\
\multirow{-2}{*}{Qwen2.5-VL-7B} & MemStream & AKS ($\times 16$) & Ensemble & 50.22 & \textbf{43.89} & 56.70 & \textbf{44.28} & \textbf{36.21} & \textbf{45.00} & \textbf{48.10} \\
\bottomrule
\end{tabular}
}
\end{table*}

\subsection{VideoMME Performance}

In the main paper, we observe that VideoMME performance sometimes experiences minimal improvement and even a slight drop. To understand this further, we investigate performance on the LLaVA-OV and InternVL3 backbones. Table~\ref{tab:videomme_breakdown} breaks down VideoMME by task type for InternVL3-8B, where MemStream matches ReKV in aggregate (Table~\ref{tab:main_results}). 
MemStream improves on tasks that require localizing specific events (action and object recognition and reasoning), but drops on tasks that require integrating information across the whole video (information synopsis and temporal reasoning). 
We attribute this to AKS retrievals concentrating on query-specific moments, which improves localization at the cost of broad coverage. LLaVA-OV-7B, where MemStream is likewise flat in aggregate, shows a similar pattern: gains on action and object recognition and losses or matching performance on temporal reasoning and information synopsis.

\noindent\begin{minipage}{\linewidth}
\begin{center}
\captionof{table}{\textbf{VideoMME Breakdown.} Per-task accuracy on the VideoMME Long subset for LLaVA-OV-7B and InternVL3-8B with ReKV and MemStream. The number of questions per task is in parentheses.}
\label{tab:videomme_breakdown}
\resizebox{0.62\linewidth}{!}{
\begin{tabular}{l cc cc}
\toprule
& \multicolumn{2}{c}{{LLaVA-OV-7B}} & \multicolumn{2}{c}{InternVL3-8B} \\
\cmidrule(lr){2-3} \cmidrule(lr){4-5}
Task Type & ReKV & MemStream & ReKV & MemStream \\
\midrule
Object Reasoning (240) & {44.6} & {\textbf{45.0}} & 55.0 & \textbf{56.2} \\
Action Reasoning (180) & {\textbf{47.2}} & {45.0} & 55.6 & \textbf{56.7} \\
Information Synopsis (163) & {67.5} & {67.5} & \textbf{73.0} & 66.9 \\
Temporal Reasoning (91) & {\textbf{37.4}} & {31.9} & \textbf{35.2} & 30.8 \\
Action Recognition (63) & {30.2} & {\textbf{38.1}} & 41.3 & \textbf{50.8} \\
Object Recognition (54) & {44.4} & {\textbf{48.1}} & 53.7 & \textbf{57.4} \\
\bottomrule
\end{tabular}
}
\end{center}
\end{minipage}

\subsection{Results on Qwen3-VL-8B}
\label{subsec:qwen3vl}

In this section, we investigate how AKS interacts with the newer Qwen3-VL-8B backbone. Table~\ref{tab:qwen3vl_token_scaling} shows that scaling the token budget still degrades recall, from 39.46 to 36.80, although more weakly than on Qwen2.5-VL (Section~\ref{sec:Analysis}) and without a corresponding drop in accuracy.

Table~\ref{tab:qwen3vl_compression} varies the AKS compression rate at ${\sim}512$ tokens per frame. Unlike Qwen2.5-VL, where compressing the sliding window by $16\times$ improves accuracy (Table~\ref{tab:token_scaling}), Qwen3-VL-8B is on par with full encoding at ${\sim}1.33\times$ and drops below it at ${\sim}4\times$, even though recall continues to improve. This mirrors our RVS-Movie analysis (Table~\ref{tab:rvs_movie_2x2}): AKS retrievals remain beneficial, but over-compressed cache representations hurt answering. It is also consistent with the much higher internal recall of Qwen3-VL-8B (37.97 vs.\ 16.27 for Qwen2.5-VL-7B, Table~\ref{tab:backbone_recall}): the cache is far less redundant, so there is little for AKS to remove. We therefore use the ${\sim}1.33\times$ setting for the Qwen3-VL-8B rows in Table~\ref{tab:main_results}.

\begin{figure*}[h!]
\begin{center}
\includegraphics[width=0.8\linewidth]{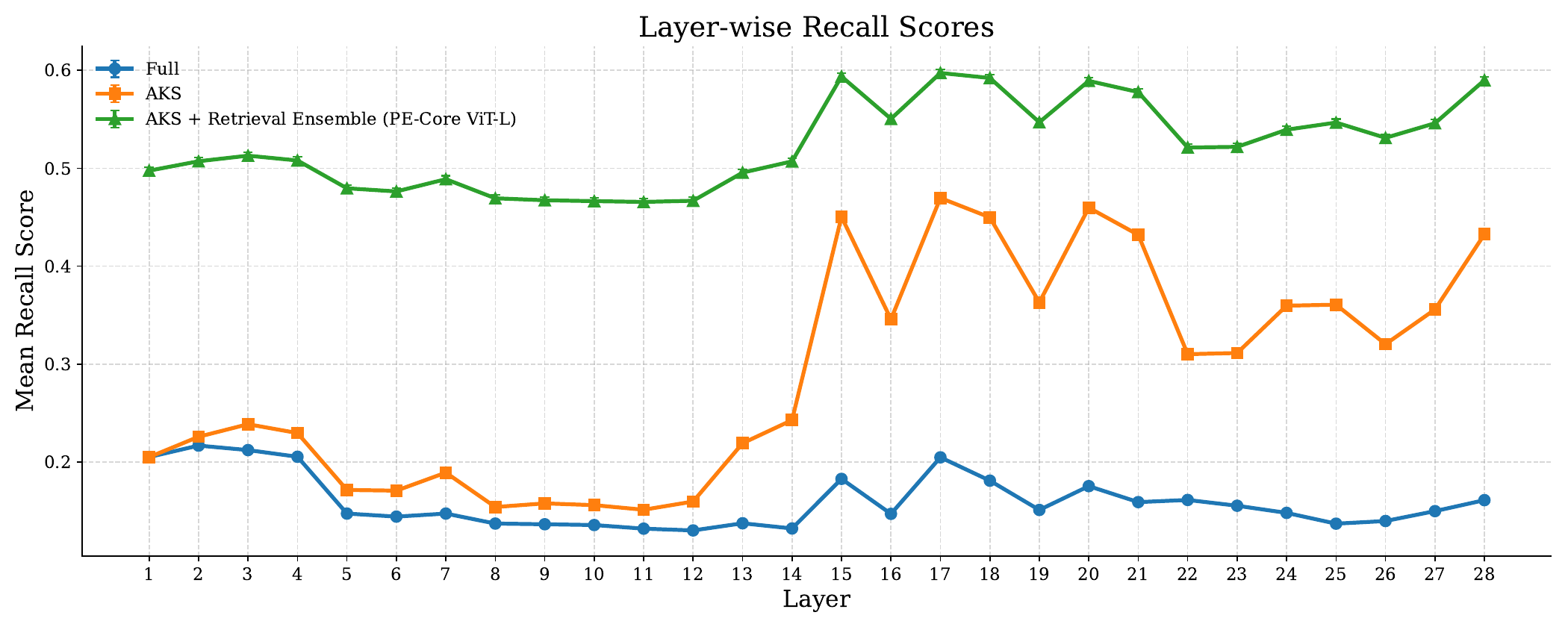}
\caption{We plot layer-wise average recall scores for full sliding-window attention, AKS, and AKS combined with retrieval ensemble with $\text{PE}_{\text{core}}$ as the external retriever.}
\label{fig:average_layerwise_retrieval}
\end{center}
\end{figure*}

To understand the drop in performance at ${\sim}4\times$ compression, we restrict AKS to the first 12 layers and observe that this retains almost all of the recall gain (39.51 vs.\ 39.68) while matching full-encoding accuracy (52.99 vs.\ 52.84), indicating that the accuracy loss comes from compressing later layers, whose representations are less redundant, while early layers can be compressed safely. This is consistent with retrieval signal concentrating in later layers (Figure~\ref{fig:average_layerwise_retrieval}). This motivates future work on adaptive compression rates. 

\begin{table}[h!]
\centering
\caption{\textbf{Token Scaling on Qwen3-VL-8B.} CG-Bench Recall@64 with full encoding (no compression) as the per-frame token budget grows.}
\label{tab:qwen3vl_token_scaling}
\resizebox{0.5\linewidth}{!}{
\begin{tabular}{lc}
\toprule
Tokens/Frame & CG-Bench Recall@64 \\
\midrule
144--160 & \textbf{39.46} \\
482--536 & 37.97 \\
700--768 & 36.80 \\
\bottomrule
\end{tabular}
}
\end{table}

\begin{table}[h!]
\centering
\caption{\textbf{AKS Compression Rate on Qwen3-VL-8B.} With a less redundant cache, mild compression matches full encoding and higher rates degrade accuracy. All rows use internal retrieval at ${\sim}512$ tokens per frame.}
\label{tab:qwen3vl_compression}
\resizebox{0.55\linewidth}{!}{
\begin{tabular}{lcccc}
\toprule
Encoding & Comp. Rate & CG-Bench Recall & CG-Bench & LVBench \\
\midrule
Full & $1\times$ & 37.97 & 52.84 & 51.45 \\
AKS  & ${\sim}1.33\times$ & 38.67 & \textbf{53.10} & \textbf{51.58} \\
AKS  & ${\sim}4\times$ & \textbf{39.68} & 52.39 & 50.55 \\
\bottomrule
\end{tabular}
}
\end{table}

\begin{table}[h!]
\centering
\caption{\textbf{Layer-Banded AKS on Qwen3-VL-8B.} Restricting ${\sim}4\times$ AKS to the first 12 layers keeps almost all of the recall gain without the accuracy loss of all-layer compression. All rows use internal retrieval at ${\sim}512$ tokens per frame.}
\label{tab:qwen3vl_banded}
\resizebox{0.6\linewidth}{!}{
\begin{tabular}{lccccc}
\toprule
Encoding & Comp. Rate & Layers & CG-Bench Recall & CG-Bench & LVBench \\
\midrule
Full & $1\times$ & -- & 37.97 & 52.84 & 51.45 \\
AKS  & ${\sim}1.33\times$ & All & 38.67 & \textbf{53.10} & 51.58 \\
AKS  & ${\sim}4\times$ & All & \textbf{39.68} & 52.39 & 50.55 \\
AKS  & ${\sim}4\times$ & 0--11 & 39.51 & 52.99 & \textbf{51.65} \\
\bottomrule
\end{tabular}
}
\end{table}

\subsection{Analysis of Layer-Wise Retrievals}
\label{subsec:layerwise_retrievals}

We plot average layer-wise recall scores for full, AKS, and AKS + Ensemble using the CG-Bench dataset in Figure~\ref{fig:average_layerwise_retrieval}.
AKS significantly improves layer-wise recall for later layers, which is further improved by our retrieval ensemble strategy.

\subsection{More Visualizations}

\begin{figure*}[t]
\begin{center}
\includegraphics[width=0.95\linewidth]{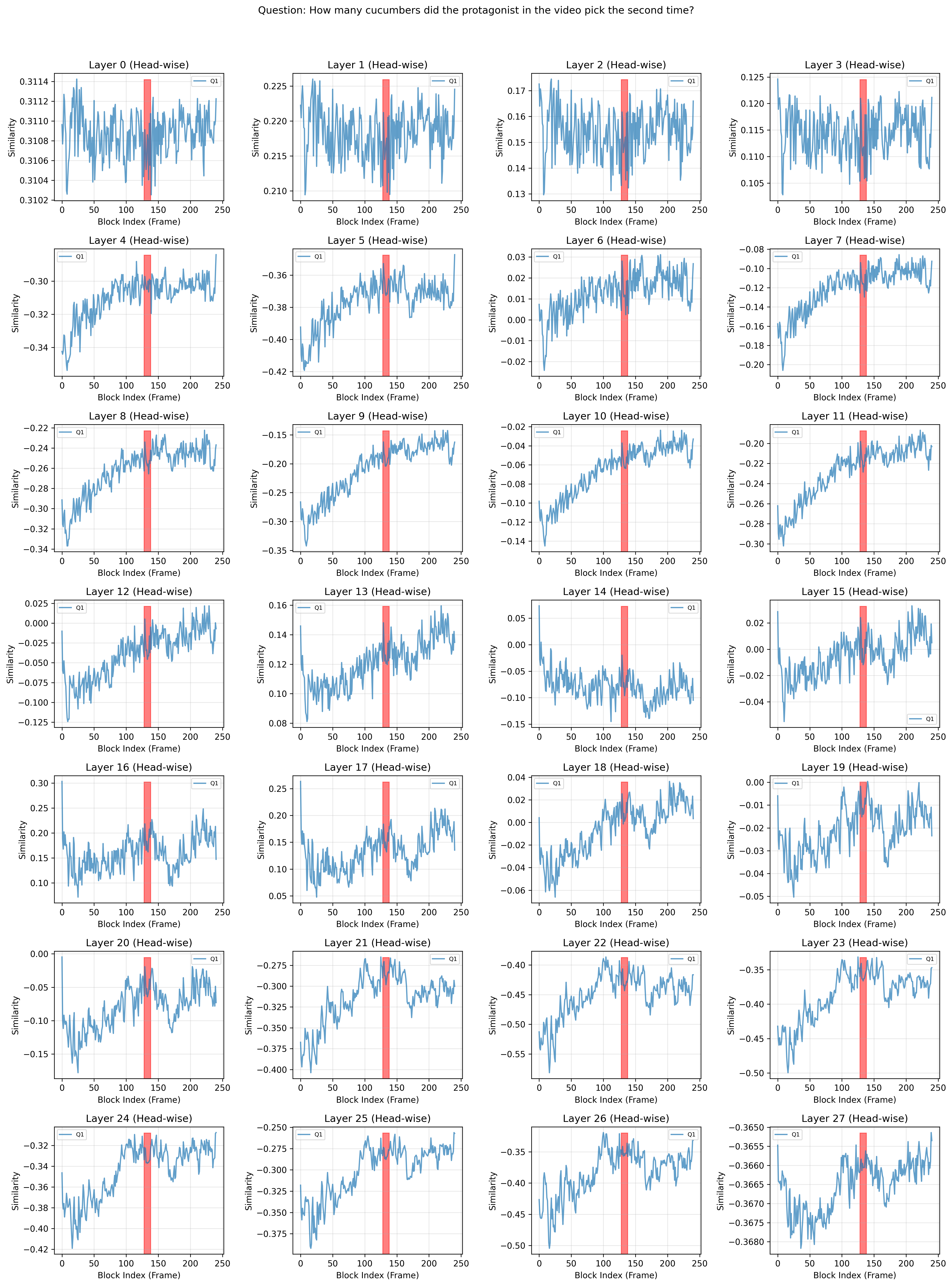}
\caption{Layer-Wise Query-Frame Scores 64 Tokens Per Frame @ 241 Frames}
\label{fig:tpf_64_scores}
\end{center}
\end{figure*}

\begin{figure*}[t]
\begin{center}
\includegraphics[width=0.95\linewidth]{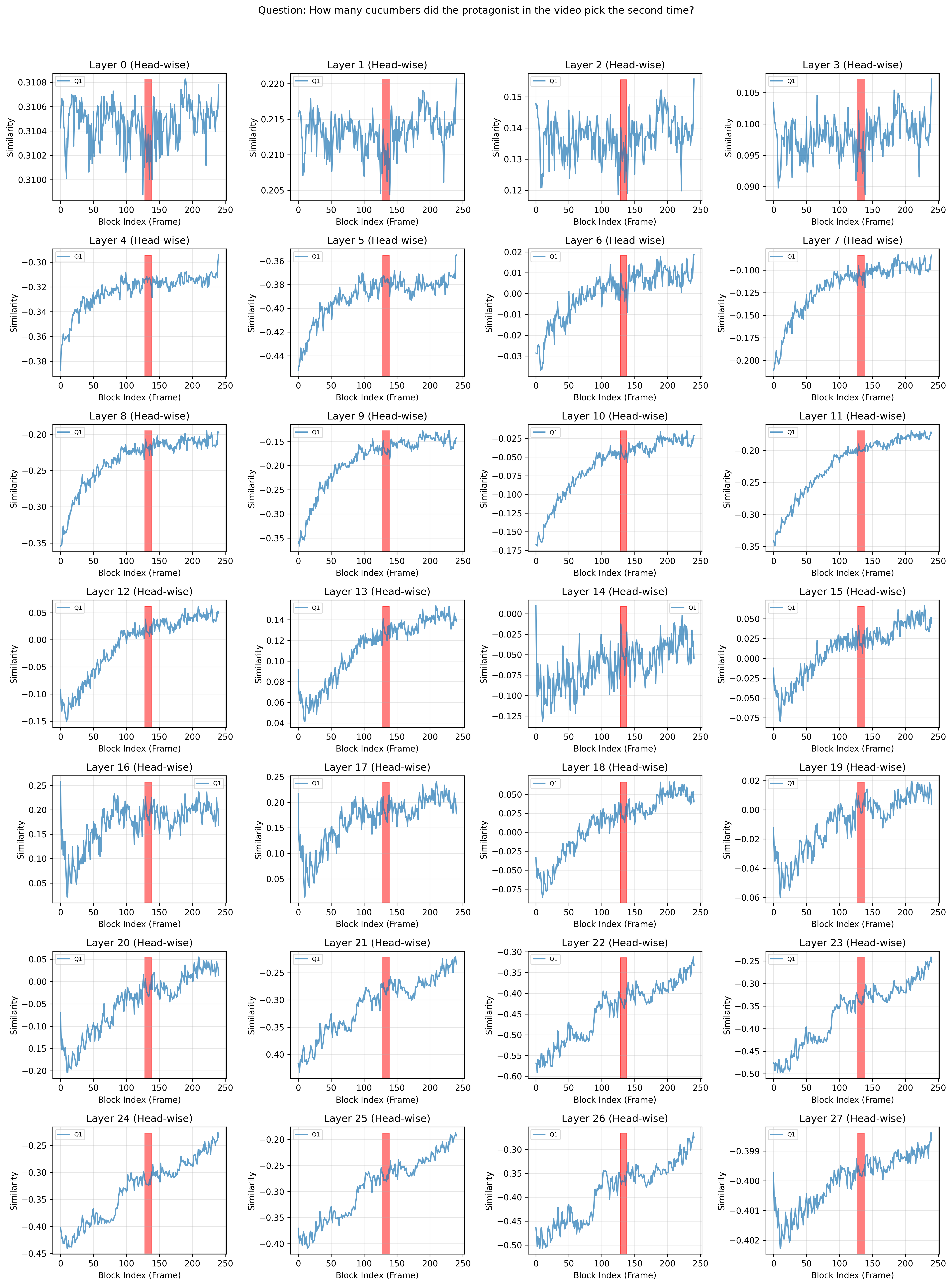}
\caption{Layer-Wise Query-Frame Scores 256 Tokens Per Frame @ 241 Frames}
\label{fig:tpf_256_scores}
\end{center}
\end{figure*}

We show layer-wise query-frame scores at 64 tokens per frame in Fig~\ref{fig:tpf_64_scores} and 256 tokens per frame in Fig~\ref{fig:tpf_256_scores}. The ground-truth segment is highlighted in red.

\begin{figure*}[t]
\begin{center}
\includegraphics[width=\linewidth]{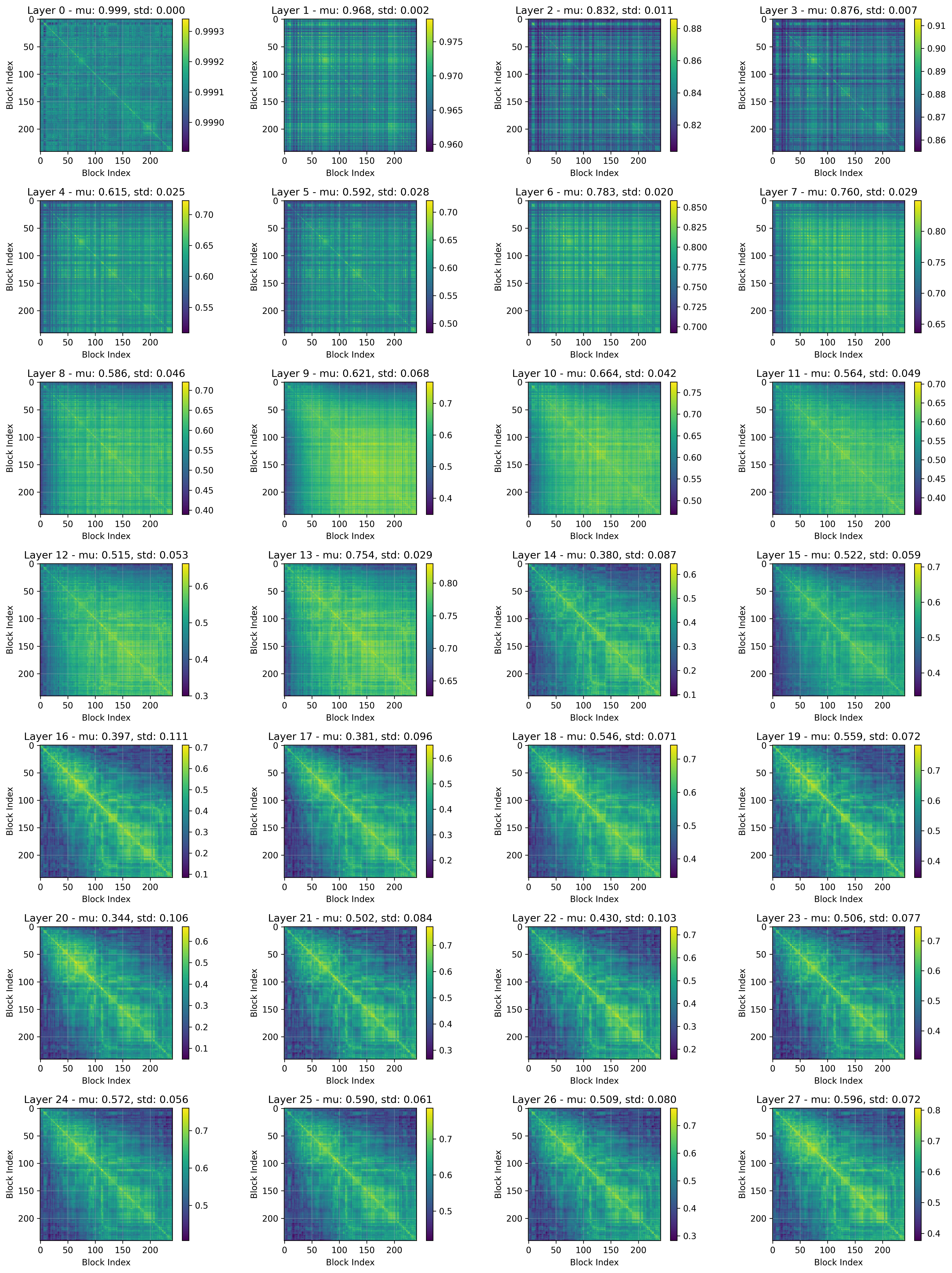}
\caption{Layer-Wise Self-Similarity Matrices at 64 Tokens Per Frame @ 241 Frames}
\label{fig:tpf_64_block}
\end{center}
\end{figure*}

\begin{figure*}[t]
\begin{center}
\includegraphics[width=\linewidth]{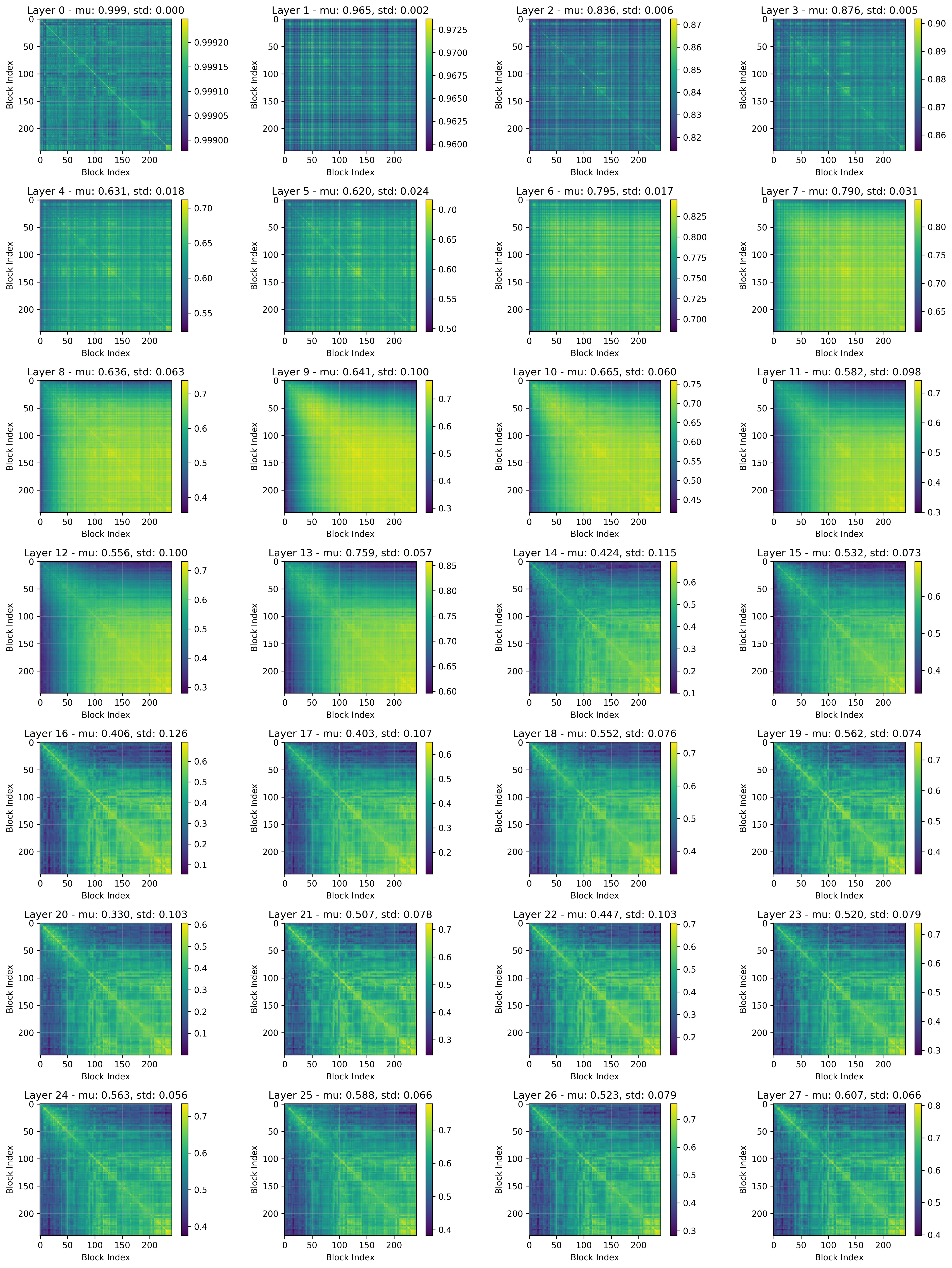}
\caption{Layer-Wise Self-Similarity Matrices at 256 Tokens Per Frame @ 241 Frames}
\label{fig:tpf_256_block}
\end{center}
\end{figure*}

We show layer-wise self-similarity matrices at 64 tokens per frame in Fig~\ref{fig:tpf_64_block} and 256 tokens per frame in Fig~\ref{fig:tpf_256_block}. We observe that at 256 tokens per frame, there is higher redundancy in the self-similarity matrices across all layers.

\section{Discussion}

\subsection{Broader Impacts} 
\label{subsec:broader_imapcts}
Generally, the use of Video-LLMs has potential risks with respect to hallucinating content, and as such, we emphasize caution when employing such models in real-world scenarios. The streaming setting has seen recent applications for smart glasses and surveillance. Although our method does not directly store images of persons or objects, we acknowledge potential privacy concerns. 

\clearpage
\section*{NeurIPS Paper Checklist}

\begin{enumerate}

\item {\bf Claims}
    \item[] Question: Do the main claims made in the abstract and introduction accurately reflect the paper's contributions and scope?
    \item[] Answer: \answerYes{} %
    \item[] Justification: Our abstract and introduction clearly reflect the paper's contributions and scope.
    \item[] Guidelines:
    \begin{itemize}
        \item The answer \answerNA{} means that the abstract and introduction do not include the claims made in the paper.
        \item The abstract and/or introduction should clearly state the claims made, including the contributions made in the paper and important assumptions and limitations. A \answerNo{} or \answerNA{} answer to this question will not be perceived well by the reviewers. 
        \item The claims made should match theoretical and experimental results, and reflect how much the results can be expected to generalize to other settings. 
        \item It is fine to include aspirational goals as motivation as long as it is clear that these goals are not attained by the paper. 
    \end{itemize}

\item {\bf Limitations}
    \item[] Question: Does the paper discuss the limitations of the work performed by the authors?
    \item[] Answer: \answerYes{} %
    \item[] Justification: The main paper discusses the limitations of the work. (Section~\ref{subsec:limitations})
    \item[] Guidelines:
    \begin{itemize}
        \item The answer \answerNA{} means that the paper has no limitation while the answer \answerNo{} means that the paper has limitations, but those are not discussed in the paper. 
        \item The authors are encouraged to create a separate ``Limitations'' section in their paper.
        \item The paper should point out any strong assumptions and how robust the results are to violations of these assumptions (e.g., independence assumptions, noiseless settings, model well-specification, asymptotic approximations only holding locally). The authors should reflect on how these assumptions might be violated in practice and what the implications would be.
        \item The authors should reflect on the scope of the claims made, e.g., if the approach was only tested on a few datasets or with a few runs. In general, empirical results often depend on implicit assumptions, which should be articulated.
        \item The authors should reflect on the factors that influence the performance of the approach. For example, a facial recognition algorithm may perform poorly when image resolution is low or images are taken in low lighting. Or a speech-to-text system might not be used reliably to provide closed captions for online lectures because it fails to handle technical jargon.
        \item The authors should discuss the computational efficiency of the proposed algorithms and how they scale with dataset size.
        \item If applicable, the authors should discuss possible limitations of their approach to address problems of privacy and fairness.
        \item While the authors might fear that complete honesty about limitations might be used by reviewers as grounds for rejection, a worse outcome might be that reviewers discover limitations that aren't acknowledged in the paper. The authors should use their best judgment and recognize that individual actions in favor of transparency play an important role in developing norms that preserve the integrity of the community. Reviewers will be specifically instructed to not penalize honesty concerning limitations.
    \end{itemize}

\item {\bf Theory assumptions and proofs}
    \item[] Question: For each theoretical result, does the paper provide the full set of assumptions and a complete (and correct) proof?
    \item[] Answer: \answerNA{} %
    \item[] Justification: There are no theoretical results. 
    \item[] Guidelines:
    \begin{itemize}
        \item The answer \answerNA{} means that the paper does not include theoretical results. 
        \item All the theorems, formulas, and proofs in the paper should be numbered and cross-referenced.
        \item All assumptions should be clearly stated or referenced in the statement of any theorems.
        \item The proofs can either appear in the main paper or the supplemental material, but if they appear in the supplemental material, the authors are encouraged to provide a short proof sketch to provide intuition. 
        \item Inversely, any informal proof provided in the core of the paper should be complemented by formal proofs provided in appendix or supplemental material.
        \item Theorems and Lemmas that the proof relies upon should be properly referenced. 
    \end{itemize}

    \item {\bf Experimental result reproducibility}
    \item[] Question: Does the paper fully disclose all the information needed to reproduce the main experimental results of the paper to the extent that it affects the main claims and/or conclusions of the paper (regardless of whether the code and data are provided or not)?
    \item[] Answer: \answerYes{} %
    \item[] Justification: The paper does provide all information to reproduce the main experimental results. Additional details about the changes needed for adopting KV-cache encoding and retrieval can be found in the supplementary. (Section~\ref{sec:Approach} and Section~\ref{subsec:qwen_integration})
    \item[] Guidelines:
    \begin{itemize}
        \item The answer \answerNA{} means that the paper does not include experiments.
        \item If the paper includes experiments, a \answerNo{} answer to this question will not be perceived well by the reviewers: Making the paper reproducible is important, regardless of whether the code and data are provided or not.
        \item If the contribution is a dataset and\slash or model, the authors should describe the steps taken to make their results reproducible or verifiable. 
        \item Depending on the contribution, reproducibility can be accomplished in various ways. For example, if the contribution is a novel architecture, describing the architecture fully might suffice, or if the contribution is a specific model and empirical evaluation, it may be necessary to either make it possible for others to replicate the model with the same dataset, or provide access to the model. In general. releasing code and data is often one good way to accomplish this, but reproducibility can also be provided via detailed instructions for how to replicate the results, access to a hosted model (e.g., in the case of a large language model), releasing of a model checkpoint, or other means that are appropriate to the research performed.
        \item While NeurIPS does not require releasing code, the conference does require all submissions to provide some reasonable avenue for reproducibility, which may depend on the nature of the contribution. For example
        \begin{enumerate}
            \item If the contribution is primarily a new algorithm, the paper should make it clear how to reproduce that algorithm.
            \item If the contribution is primarily a new model architecture, the paper should describe the architecture clearly and fully.
            \item If the contribution is a new model (e.g., a large language model), then there should either be a way to access this model for reproducing the results or a way to reproduce the model (e.g., with an open-source dataset or instructions for how to construct the dataset).
            \item We recognize that reproducibility may be tricky in some cases, in which case authors are welcome to describe the particular way they provide for reproducibility. In the case of closed-source models, it may be that access to the model is limited in some way (e.g., to registered users), but it should be possible for other researchers to have some path to reproducing or verifying the results.
        \end{enumerate}
    \end{itemize}

\item {\bf Open access to data and code}
    \item[] Question: Does the paper provide open access to the data and code, with sufficient instructions to faithfully reproduce the main experimental results, as described in supplemental material?
    \item[] Answer: \answerNo{} %
    \item[] Justification: At this time, we are unable to provide code.
    \item[] Guidelines:
    \begin{itemize}
        \item The answer \answerNA{} means that paper does not include experiments requiring code.
        \item Please see the NeurIPS code and data submission guidelines (\url{https://neurips.cc/public/guides/CodeSubmissionPolicy}) for more details.
        \item While we encourage the release of code and data, we understand that this might not be possible, so \answerNo{} is an acceptable answer. Papers cannot be rejected simply for not including code, unless this is central to the contribution (e.g., for a new open-source benchmark).
        \item The instructions should contain the exact command and environment needed to run to reproduce the results. See the NeurIPS code and data submission guidelines (\url{https://neurips.cc/public/guides/CodeSubmissionPolicy}) for more details.
        \item The authors should provide instructions on data access and preparation, including how to access the raw data, preprocessed data, intermediate data, and generated data, etc.
        \item The authors should provide scripts to reproduce all experimental results for the new proposed method and baselines. If only a subset of experiments are reproducible, they should state which ones are omitted from the script and why.
        \item At submission time, to preserve anonymity, the authors should release anonymized versions (if applicable).
        \item Providing as much information as possible in supplemental material (appended to the paper) is recommended, but including URLs to data and code is permitted.
    \end{itemize}

\item {\bf Experimental setting/details}
    \item[] Question: Does the paper specify all the training and test details (e.g., data splits, hyperparameters, how they were chosen, type of optimizer) necessary to understand the results?
    \item[] Answer: \answerYes{} %
    \item[] Justification: Yes all test details are provided, see Section~\ref{sec:experiments}.
    \item[] Guidelines:
    \begin{itemize}
        \item The answer \answerNA{} means that the paper does not include experiments.
        \item The experimental setting should be presented in the core of the paper to a level of detail that is necessary to appreciate the results and make sense of them.
        \item The full details can be provided either with the code, in appendix, or as supplemental material.
    \end{itemize}

\item {\bf Experiment statistical significance}
    \item[] Question: Does the paper report error bars suitably and correctly defined or other appropriate information about the statistical significance of the experiments?
    \item[] Answer: \answerNo{} %
    \item[] Justification: Due to computational costs, we were unable to report error bars. 
    \item[] Guidelines:
    \begin{itemize}
        \item The answer \answerNA{} means that the paper does not include experiments.
        \item The authors should answer \answerYes{} if the results are accompanied by error bars, confidence intervals, or statistical significance tests, at least for the experiments that support the main claims of the paper.
        \item The factors of variability that the error bars are capturing should be clearly stated (for example, train/test split, initialization, random drawing of some parameter, or overall run with given experimental conditions).
        \item The method for calculating the error bars should be explained (closed form formula, call to a library function, bootstrap, etc.)
        \item The assumptions made should be given (e.g., Normally distributed errors).
        \item It should be clear whether the error bar is the standard deviation or the standard error of the mean.
        \item It is OK to report 1-sigma error bars, but one should state it. The authors should preferably report a 2-sigma error bar than state that they have a 96\% CI, if the hypothesis of Normality of errors is not verified.
        \item For asymmetric distributions, the authors should be careful not to show in tables or figures symmetric error bars that would yield results that are out of range (e.g., negative error rates).
        \item If error bars are reported in tables or plots, the authors should explain in the text how they were calculated and reference the corresponding figures or tables in the text.
    \end{itemize}

\item {\bf Experiments compute resources}
    \item[] Question: For each experiment, does the paper provide sufficient information on the computer resources (type of compute workers, memory, time of execution) needed to reproduce the experiments?
    \item[] Answer: \answerYes{} %
    \item[] Justification: We provide memory and latency results on an H200 GPU.  
    \item[] Guidelines:
    \begin{itemize}
        \item The answer \answerNA{} means that the paper does not include experiments.
        \item The paper should indicate the type of compute workers CPU or GPU, internal cluster, or cloud provider, including relevant memory and storage.
        \item The paper should provide the amount of compute required for each of the individual experimental runs as well as estimate the total compute. 
        \item The paper should disclose whether the full research project required more compute than the experiments reported in the paper (e.g., preliminary or failed experiments that didn't make it into the paper). 
    \end{itemize}
    
\item {\bf Code of ethics}
    \item[] Question: Does the research conducted in the paper conform, in every respect, with the NeurIPS Code of Ethics \url{https://neurips.cc/public/EthicsGuidelines}?
    \item[] Answer: \answerYes{} %
    \item[] Justification:  We have read the NeurIPS Code of Ethics and declare no deviation from it. We address concerns with Video-LLMs here, namely that these models are still in development and should be used cautiously. 
    \item[] Guidelines:
    \begin{itemize}
        \item The answer \answerNA{} means that the authors have not reviewed the NeurIPS Code of Ethics.
        \item If the authors answer \answerNo, they should explain the special circumstances that require a deviation from the Code of Ethics.
        \item The authors should make sure to preserve anonymity (e.g., if there is a special consideration due to laws or regulations in their jurisdiction).
    \end{itemize}

\item {\bf Broader impacts}
    \item[] Question: Does the paper discuss both potential positive societal impacts and negative societal impacts of the work performed?
    \item[] Answer: \answerYes{} %
    \item[] Justification: Yes, these are discussed in Section~\ref{subsec:broader_imapcts}.
    \item[] Guidelines:
    \begin{itemize}
        \item The answer \answerNA{} means that there is no societal impact of the work performed.
        \item If the authors answer \answerNA{} or \answerNo, they should explain why their work has no societal impact or why the paper does not address societal impact.
        \item Examples of negative societal impacts include potential malicious or unintended uses (e.g., disinformation, generating fake profiles, surveillance), fairness considerations (e.g., deployment of technologies that could make decisions that unfairly impact specific groups), privacy considerations, and security considerations.
        \item The conference expects that many papers will be foundational research and not tied to particular applications, let alone deployments. However, if there is a direct path to any negative applications, the authors should point it out. For example, it is legitimate to point out that an improvement in the quality of generative models could be used to generate Deepfakes for disinformation. On the other hand, it is not needed to point out that a generic algorithm for optimizing neural networks could enable people to train models that generate Deepfakes faster.
        \item The authors should consider possible harms that could arise when the technology is being used as intended and functioning correctly, harms that could arise when the technology is being used as intended but gives incorrect results, and harms following from (intentional or unintentional) misuse of the technology.
        \item If there are negative societal impacts, the authors could also discuss possible mitigation strategies (e.g., gated release of models, providing defenses in addition to attacks, mechanisms for monitoring misuse, mechanisms to monitor how a system learns from feedback over time, improving the efficiency and accessibility of ML).
    \end{itemize}
    
\item {\bf Safeguards}
    \item[] Question: Does the paper describe safeguards that have been put in place for responsible release of data or models that have a high risk for misuse (e.g., pre-trained language models, image generators, or scraped datasets)?
    \item[] Answer: \answerNA{}{} %
    \item[] Justification: We are not releasing any data or models. 
    \item[] Guidelines:
    \begin{itemize}
        \item The answer \answerNA{} means that the paper poses no such risks.
        \item Released models that have a high risk for misuse or dual-use should be released with necessary safeguards to allow for controlled use of the model, for example by requiring that users adhere to usage guidelines or restrictions to access the model or implementing safety filters. 
        \item Datasets that have been scraped from the Internet could pose safety risks. The authors should describe how they avoided releasing unsafe images.
        \item We recognize that providing effective safeguards is challenging, and many papers do not require this, but we encourage authors to take this into account and make a best faith effort.
    \end{itemize}

\item {\bf Licenses for existing assets}
    \item[] Question: Are the creators or original owners of assets (e.g., code, data, models), used in the paper, properly credited and are the license and terms of use explicitly mentioned and properly respected?
    \item[] Answer: \answerYes{} %
    \item[] Justification: We cite all owners of assets used in this paper, including models and datasets. 
    \item[] Guidelines:
    \begin{itemize}
        \item The answer \answerNA{} means that the paper does not use existing assets.
        \item The authors should cite the original paper that produced the code package or dataset.
        \item The authors should state which version of the asset is used and, if possible, include a URL.
        \item The name of the license (e.g., CC-BY 4.0) should be included for each asset.
        \item For scraped data from a particular source (e.g., website), the copyright and terms of service of that source should be provided.
        \item If assets are released, the license, copyright information, and terms of use in the package should be provided. For popular datasets, \url{paperswithcode.com/datasets} has curated licenses for some datasets. Their licensing guide can help determine the license of a dataset.
        \item For existing datasets that are re-packaged, both the original license and the license of the derived asset (if it has changed) should be provided.
        \item If this information is not available online, the authors are encouraged to reach out to the asset's creators.
    \end{itemize}

\item {\bf New assets}
    \item[] Question: Are new assets introduced in the paper well documented and is the documentation provided alongside the assets?
    \item[] Answer: \answerNA{} %
    \item[] Justification: We do not introduce any new assets in this paper. 
    \item[] Guidelines:
    \begin{itemize}
        \item The answer \answerNA{} means that the paper does not release new assets.
        \item Researchers should communicate the details of the dataset\slash code\slash model as part of their submissions via structured templates. This includes details about training, license, limitations, etc. 
        \item The paper should discuss whether and how consent was obtained from people whose asset is used.
        \item At submission time, remember to anonymize your assets (if applicable). You can either create an anonymized URL or include an anonymized zip file.
    \end{itemize}

\item {\bf Crowdsourcing and research with human subjects}
    \item[] Question: For crowdsourcing experiments and research with human subjects, does the paper include the full text of instructions given to participants and screenshots, if applicable, as well as details about compensation (if any)? 
    \item[] Answer: \answerNA{} %
    \item[] Justification: We do not involve crowdsourcing or research with human subjects. 
    \item[] Guidelines:
    \begin{itemize}
        \item The answer \answerNA{} means that the paper does not involve crowdsourcing nor research with human subjects.
        \item Including this information in the supplemental material is fine, but if the main contribution of the paper involves human subjects, then as much detail as possible should be included in the main paper. 
        \item According to the NeurIPS Code of Ethics, workers involved in data collection, curation, or other labor should be paid at least the minimum wage in the country of the data collector. 
    \end{itemize}

\item {\bf Institutional review board (IRB) approvals or equivalent for research with human subjects}
    \item[] Question: Does the paper describe potential risks incurred by study participants, whether such risks were disclosed to the subjects, and whether Institutional Review Board (IRB) approvals (or an equivalent approval/review based on the requirements of your country or institution) were obtained?
    \item[] Answer: \answerNA{} %
    \item[] Justification: No crowdsourcing nor research with human subjects was conducted. 
    \item[] Guidelines:
    \begin{itemize}
        \item The answer \answerNA{} means that the paper does not involve crowdsourcing nor research with human subjects.
        \item Depending on the country in which research is conducted, IRB approval (or equivalent) may be required for any human subjects research. If you obtained IRB approval, you should clearly state this in the paper. 
        \item We recognize that the procedures for this may vary significantly between institutions and locations, and we expect authors to adhere to the NeurIPS Code of Ethics and the guidelines for their institution. 
        \item For initial submissions, do not include any information that would break anonymity (if applicable), such as the institution conducting the review.
    \end{itemize}

\item {\bf Declaration of LLM usage}
    \item[] Question: Does the paper describe the usage of LLMs if it is an important, original, or non-standard component of the core methods in this research? Note that if the LLM is used only for writing, editing, or formatting purposes and does \emph{not} impact the core methodology, scientific rigor, or originality of the research, declaration is not required.
    \item[] Answer: \answerNA{} %
    \item[] Justification: The core development in this research does not involve any LLMs. 
    \item[] Guidelines:
    \begin{itemize}
        \item The answer \answerNA{} means that the core method development in this research does not involve LLMs as any important, original, or non-standard components.
        \item Please refer to our LLM policy in the NeurIPS handbook for what should or should not be described.
    \end{itemize}

\end{enumerate}

\end{document}